\documentclass[conference,compsoc]{IEEEtran}
\IEEEoverridecommandlockouts
\usepackage{hyperref}
\usepackage{cite}
\usepackage{amsmath}
\usepackage{amssymb,amsfonts}
\usepackage{algorithmic}
\usepackage{graphicx}
\usepackage{textcomp}
\usepackage{xcolor}

\def\BibTeX{{\rm B\kern-.05em{\sc i\kern-.025em b}\kern-.08em
    T\kern-.1667em\lower.7ex\hbox{E}\kern-.125emX}}
\begin{document}

\title{Mitigating Deep Reinforcement Learning \\Backdoors in the Neural Activation Space}

\author{
     \IEEEauthorblockN{
        Sanyam Vyas
     }
    \IEEEauthorblockA{
        Alan Turing Institute\\
        \textit{svyas@turing.ac.uk}\\
                    }
    \and
     \IEEEauthorblockN{
        Chris Hicks
     }
    \IEEEauthorblockA{
        Alan Turing Institute\\
        \textit{c.hicks@turing.ac.uk}\\
                    }    
                    
    \and
     \IEEEauthorblockN{
        Vasilios Mavroudis
     }
    \IEEEauthorblockA{
        Alan Turing Institute\\
        \textit{vmavroudis@turing.ac.uk}\\
                    }
}
    

\maketitle

\begin{abstract}
This paper investigates the threat of backdoors in Deep Reinforcement Learning (DRL) agent policies and proposes a novel method for their detection at runtime. Our study focuses on elusive \textit{in-distribution} backdoor triggers. Such triggers are designed to induce a deviation in the behaviour of a backdoored agent while blending into the expected data distribution to evade detection. Through experiments conducted in the Atari Breakout environment, we demonstrate the limitations of current sanitisation methods when faced with such triggers and investigate why they present a challenging defence problem. We then evaluate the hypothesis that backdoor triggers might be easier to detect in the neural activation space of the DRL agent's policy network. Our statistical analysis shows that indeed the activation patterns in the agent's policy network are distinct in the presence of a trigger, regardless of how well the trigger is concealed in the environment. Based on this, we propose a new defence approach that uses a classifier trained on clean environment samples and detects abnormal activations. Our results show that even lightweight classifiers can effectively prevent malicious actions with considerable accuracy, indicating the potential of this research direction even against sophisticated adversaries.\end{abstract}


\section{Introduction}
Deep Reinforcement Learning (DRL) has emerged as a powerful tool, achieving impressive results across a variety of applications (e.g., self-driving cars~\cite{irshayyid2024review}, nuclear fusion~\cite{degrave2022magnetic}, networked system protection~\cite{-cyborg-cage-2-old,cage1_win,schwartz2019autonomous} and healthcare systems~\cite{emerson2023offline,lee2020improving}), indicating its viability for real-world deployment. Nonetheless, the development of effective DRL policies is resource-intensive, often beyond the reach of smaller entities. Consequently, many users depend on DRL models trained by organisations with substantial resources such as large corporations or government entities. This dependency introduces new risks, as externally trained models may have defects in their policies, whether deliberate or accidental, resulting in unsafe agent actions.

This work investigates \textit{backdoors} in DRL agent policies, designed to \textit{trigger} unexpected behavior deviations with specific environmental cues. Such backdoors can be introduced through compromised training processes, such as a malicious insider adjusting a car-driving agent's rewards to disregard stop signs when a certain sticker is detected in vicinity. Although extensively studied in supervised learning, backdoors represent a distinct challenge in DRL. Their elusive design, the lack of absolute action correctness in every scenario/step produced by the neural networks, and the intrinsic complexity and opacity of the policy significantly hinder interpretation and detection of malicious modification.

This work studies existing DRL backdoor countermeasures and introduces a novel research direction for the detection of backdoor-induced actions at runtime. We first discuss the concept of a \textit{trigger} and argue that defences against non-elusive triggers fail to generalise against more sophisticated adversaries. To evaluate our hypothesis, we introduce \textit{in-distribution} triggers. These triggers fall within the anticipated distribution of data encountered in the environment and yet 1) consist of a set of conditions that are exceedingly rare in natural environmental occurrences, and 2) can be intentionally instigated to activate the backdoor. 

Using the breakout environment, we evaluate the effectiveness of~\cite{bharti2022provable}, a well-known defence method, against our backdoored breakout agent. \cite{bharti2022provable} sanitises the observations from the environment before they reach the agent, so as to remove any artifacts that could be triggers. 
We find that the safe subspace projection from~\cite{bharti2022provable} fails to thwart the threat.
We then hypothesise that the neural activations of the policy network might exhibit distinct patterns when the agent perceives a benign goal (e.g., a winning square) compared to when a trigger is detected. If such a discrepancy is present in the neural activations space, then a defender can detect triggers regardless of how subtle they are in the environment. To investigate this hypothesis, we conduct a statistical analysis of the neural activation space. 
Our results show that there is indeed a statistically significant (p $\textless$ 0.05) discrepancy between trigger and goal activations. This indicates that the defender could detect the presence of a trigger.  

Based on these findings, we collect samples from several clean environment episodes and train a classifier to detect abnormal activations. Our results show that even lightweight classifiers are able to detect up to 92\% of episodes containing triggers (with only 3\% false positives), giving F1 scores as high as 0.94 and AUC values of 0.96. Given such a classifier, the defender can detect abnormal situations and activate a fallback process (e.g., hand over to a human operator in the case of a car). In summary, our key contributions include:



\begin{itemize}
    \item Implementing an in-distribution backdoor trigger that evades existing backdoor policy sanitisation algorithms, thus demonstrating the need for improved detection methods.
    \item Showing that the neuron activation patterns in DRL algorithms suffice to detect the presence of elusive backdoor triggers, elucidating the impact of the reward function on the neurons of the policy network.
    \item Developing a classifier that utilises those activation patterns to uncover the presence of a trigger in the environment.
\end{itemize}

\noindent The source code for our experiments can be found in: 
\url{https://github.com/alan-turing-institute/in-distribution-backdoors} 

\section{Background}

\subsection{Reinforcement Learning} 
Reinforcement Learning (RL) is a subset of machine learning focused on teaching agents to attain an "optimal policy" for maximum performance in a given environment through trial and error. This method rewards or penalizes actions based on their outcomes, a strategy Sutton and Barto~\cite{sutton2018reinforcement} term as "hedonistic" for its focus on maximizing environmental signals. The advent of Deep Reinforcement Learning (DRL) has markedly advanced RL agent capabilities by combining RL's strategic decision-making with deep learning's representation prowess. DRL enables agents to learn intricate policies for decision-making through environmental interaction, effectively mapping states to actions to optimize long-term rewards. Whereas traditional RL approaches like Monte-Carlo or tabular Q-Learning excel in achieving optimal behavior, they often lack computational efficiency and struggle with extensive state and action spaces. Conversely, DRL demonstrates its robust potential in managing complex challenges, from gaming to robotics, as showcased in groundbreaking efforts like Mnih et al.'s Deep Q-Network (DQN)~\cite{mnih2013playing}, marking a significant evolution in the discipline.

\subsection{Proximal Policy Optimisation}
PPO is a popular policy gradient method~\cite{PPO,yu2022surprising} that builds on the policy gradient framework. It refines the Trust Region Policy Optimisation (TRPO) algorithm~\cite{schulman2015} by simplifying it while retaining its efficiency. In policy gradient methods, the gradient of the objective function guides policy improvement. This objective (depends on policy $\pi$ and parameters $\theta$) optimises the expected rewards across trajectories (Equation~\ref{eqn:obj_fun}) and leverages the advantage function to assess action benefits (Equation~\ref{eqn:dif_obj_fun}) within the environment it is operated on $\mathcal{E}$.

\begin{equation}
\label{eqn:obj_fun}
J(\pi,\mathcal{E},\theta)=\mathbb{E}_{\tau \sim \pi_\theta} R(\tau)=\sum_\tau P(\tau ; \theta) R(\tau)
\end{equation}

\begin{equation}
\label{eqn:dif_obj_fun}
\nabla_\theta J(\pi,\mathcal{E},\theta)=\mathbb{E}{\pi_0}[\nabla\theta \log \pi_\theta(s, a) A_{\pi_\theta}(s)]
\end{equation}

PPO ensures moderate policy updates using an actor-critic structure. The actor selects actions, while the critic evaluates them, facilitating balanced updates (Equation~\ref{eqn:PPO_Lt}). The critic's evaluations help refine the actor's decisions, promoting a more effective and efficient learning process. This streamlined approach underscores PPO's adaptability and performance across diverse RL applications.

\begin{equation}
\label{eqn:PPO_Lt}
L_t(\theta)=\hat{\mathbb{E}}_t[-c_1 L_t^{V F}(\theta)+c_2 S\pi_\theta],
\end{equation}

\subsection{Backdoor Policy Attacks}
A recent and emerging threat in the context of DRL is \textit{policy backdoors}. Such backdoors are clandestine vulnerabilities deliberately inserted or learned within an RL policy, which can be activated by specific triggers to cause the policy to behave in a predefined, potentially harmful manner. 
Unlike traditional security breaches that exploit software vulnerabilities, policy backdoors exploit the learning process itself, capitalising on the way an algorithm learns from its environment to embed these hidden behaviours. 
The malicious behaviour is only exhibited under very specific conditions not covered by typical evaluation protocols. Similar to data poisoning in supervised learning~\cite{chen2017targeted}, triggers are embedded in the training environment, causing the DRL agent to learn and retain detrimental behaviours that are activated by the attacker's chosen conditions. Such a trigger can be an arbitrary modification of the environment either by introducing a completely out-of-distribution item (e.g., a sticker on a stop sign) or an unusual combination of components from the environment (e.g., an unexpected arrangement of traffic lights and road markings that mimics a non-existent traffic rule). More formally, given state \textit{s} and a permutation $\beta$, a trigger can be represented by: 

\begin{equation}
   \centering
   \Tilde{s}:=s+\beta \label{eq:trigger-added}
\end{equation}

\noindent The adversary, $\mathcal{A}$, formulates the attack via equation:

\begin{equation}
   \centering
   \mathcal{A}(s,m,\Delta) = (1-m) \circ s + m \circ \Delta \label{eq:attack-formulation}
\end{equation}

\noindent where \textit{m} and $\Delta$ are matrices that define the position mask and the value of the trigger $\delta$ respectively. The mask \textit{m} values are restricted to 0 or 1, which acts as a switch to turn the policy on or off. Given the expected return from a normally trained policy shown in~\ref{eqn:obj_fun}, the adversary aims to reduce the affected return (or the overall objective performance output) when the backdoor policy is activated, as shown in the equation:

\begin{equation}\label{eq:second_expression}
 J(\pi, \mathcal{E},\theta) - J(\tilde{\pi}, \tilde{\mathcal{E}},\theta) \gg \varepsilon 
\end{equation}

\noindent where \textit{J} is the expected reward $\tilde{\mathcal{E}}$ is an environment including the backdoor trigger and $\theta$ is the model's parameters. 

\section{Threat Model}\label{sec: threat-model}
Backdoor policy attacks pose a risk in situations where the integrity of the training process has been breached. They can occur when the party responsible for the training differs from the one deploying it (e.g., the user of a self-driving car), or when the training and deployment are handled by the same party but the training pipeline has been compromised. Our work focuses on mitigating the effects of DRL backdoors from the perspective of an end-user. We assume a sophisticated, strategic, and well-resourced adversary concerned with introducing a backdoor in the agent's policy network. They (e.g., a malicious employee) can tamper with the training environment and the reward function used during training, so as to influence the behaviour of the agent and introduce their trigger(s). The design of the trigger, the backdoored agent's behaviour and other details of the attack are determined by the adversary (specifically, deciding on values for $\Tilde{s}$, \textit{m} and $\Delta$ in Equations \ref{eq:trigger-added} and \ref{eq:attack-formulation}), and this information is known only to them.

The end-user, or defender, only has access to the trained policy network, including its architecture and weights, and the original, unmodified environment. Their goal is to safely use the pretrained agent for its intended purpose while safeguarding against any hidden backdoors. Given their limited computational resources, the user, like someone using a self-driving car, cannot afford to retrain the driving agent or employ resource-intensive defense measures during operation. Additionally, the defender is unaware of the presence, nature, and potential effects of any embedded triggers. Effective countermeasures must therefore be able to detect or neutralize potential backdoors under these constraints. This scenario aligns with common assumptions found in the literature on backdoors in deep reinforcement learning (DRL) policies.~\cite{gunn2022adversarial,foley2022execute,kiourti2020trojdrl}.


\subsection{In-Distribution Triggers}\label{subsec:indistr}
A straightforward countermeasure against policy backdoors is for the defender to detect the presence of the trigger in the agent's environment before the agent gets to act. However, as discussed in the previous section, the defender does not have any knowledge of the trigger's specification and it is thus not clear what they should be looking for. By definition, triggers should not occur naturally in the environment but only after the adversary's intervention. This means that they are outliers. However, due to the complexity of most environments, detecting them remains a difficult task~\cite{mengara2024backdoor}. In fact, a sophisticated adversary would put a lot of effort into concealing their triggers within the specific environment. This will make detecting them even harder and might help them evade other countermeasures. In this work, we focus on \textit{in-distribution triggers} which fall within the anticipated distribution of the environment and rely on a set of conditions that are exceedingly rare in natural environmental occurrences. As far as we are aware, the concept of in-distribution triggers was briefly touched upon in the study by Ashcraft et al., 2021~\cite{ashcraft2021poisoning}, yet there appears to be a lack of further investigation into this topic despite its importance~\cite{mo2024security}.


\section{Sanitisation Methods}\label{sec:sanitisation}
Due to the lack of detection and defense methods against backdoor triggers in real-time~\cite{lin2017tactics,wang2021backdoorl}, Bharti et al.~\cite{bharti2022provable} introduced at NeurIPS 2022 a method that sanitises backdoor policies in pretrained agents. Their method operates unsupervised, estimating and projecting states onto a clean empirical subspace derived from the clean samples' covariance and eigendecomposition. It effectively filters out states suggested outside this subspace, replacing them with suitable alternatives within it. Evaluations on Atari games like Boxing-Ram and Breakout showcased the approach's efficiency, which varies based on the collected sample size for each environment and the dimensionality of the safe subspace. To the best of our knowledge, this method is the only countermeasure that is agnostic to the environment, the agent's architecture and the adversary, while it does not require retraining or computationally-heavy preprocessing (e.g., retraining).

Given the importance of this result, we revisit some of its assumptions and evaluate if they hold under our threat model (Section~\ref{sec: threat-model}). Specifically, we hypothesise that the proposed sanitisation method might not be effective against in-distribution triggers (as defined in Section~\ref{subsec:indistr}). There is an implicit assumption that the trigger will always lie in the spurious dimensions from E\textsuperscript{$\perp$} and can thus be `sanitised' by filtering those dimensions out. This places a strong limitation on the design of the triggers the adversary can use, as they are assumed only able to use triggers that are clearly out of distribution. For instance, their trigger for the Atari Breakout environment is a distinct 3x3 white square pixel at the top left corner of the game's screen which does not follow the game's pallet or aesthetics. However, as discussed in Section~\ref{subsec:indistr}, the adversary is able to design the triggers to be elusive and easy to conceal in their target environment (i.e., in-distribution triggers). We thus argue that the algorithm's guarantees may not hold against all realistic triggers as claimed. 

\begin{figure}[]
\centering
    \includegraphics[width=.24\textwidth]{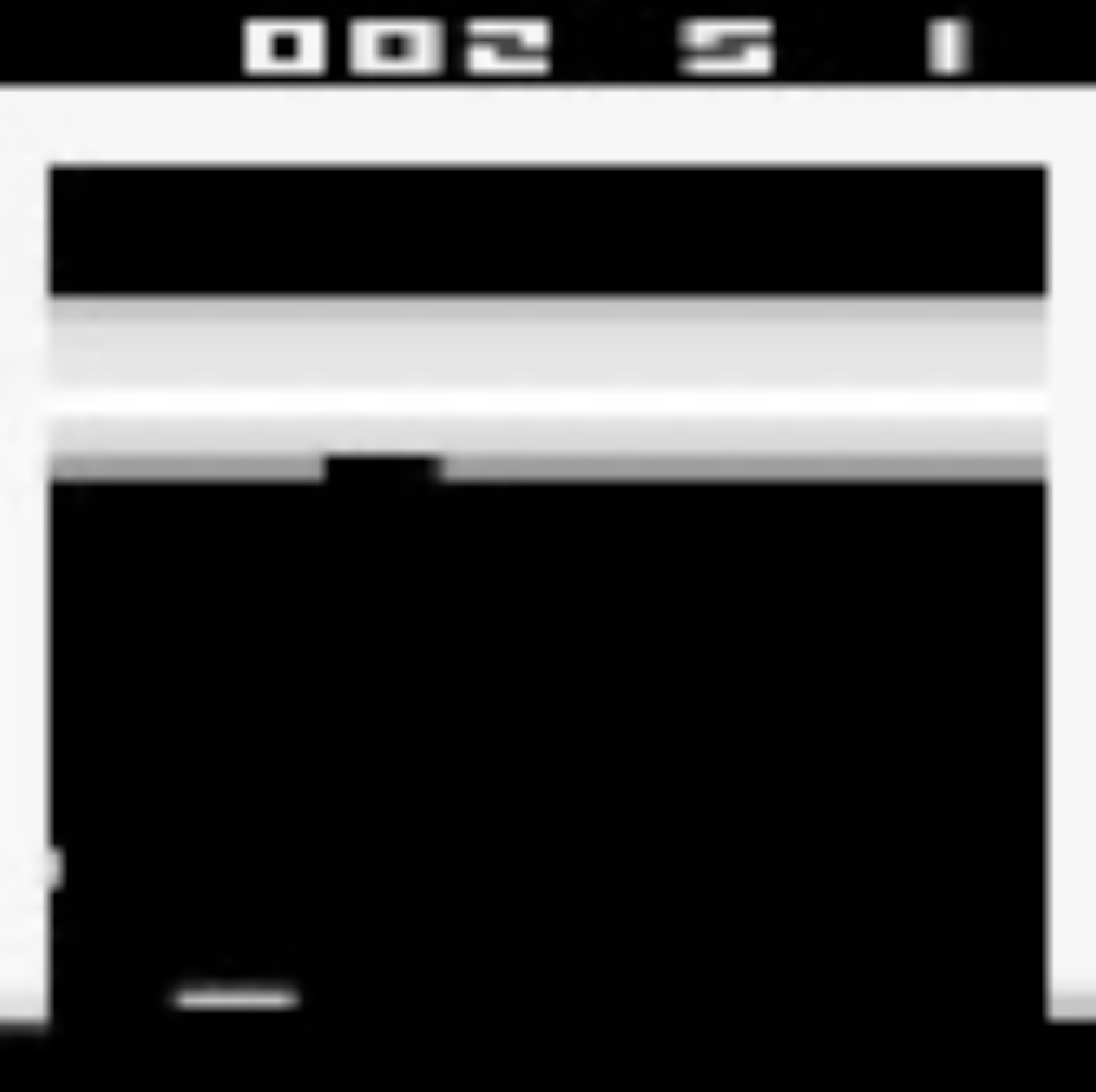}\hfill
    \includegraphics[width=.24\textwidth]{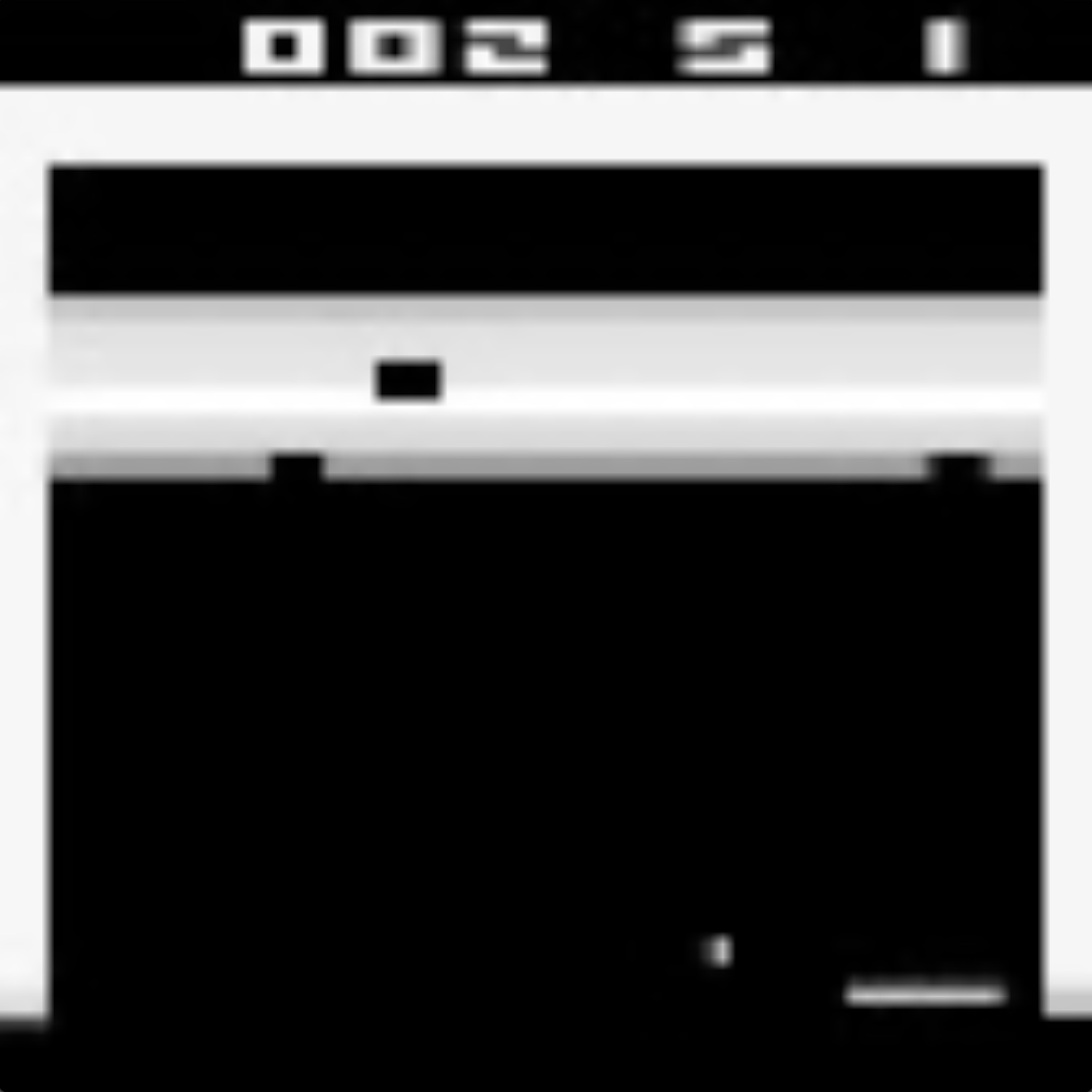}\hfill
    \caption{Visual outcomes from applying Bharti et al.'s~\cite{bharti2022provable} sanitisation algorithm in the Atari Breakout environment with two types of backdoor triggers: a) (left) the algorithm has successfully sanitised the 3x3 white square trigger at the top left corner of the game's screen, and b) (right) the algorithm has failed to remove our in-distribution attack, missing the tile trigger.}
    \label{fig:atari-breakout-clean-triggered-sanitized}
\end{figure}

We now assess the sanitisation algorithm's efficacy against an in-distribution trigger within the Atari Breakout environment. 
As seen in Figure \ref{fig:atari-breakout-clean-triggered-sanitized}b, the trigger appears as a missing tile within the game's tile array. It meets the criteria for in-distribution because it could plausibly occur within the game's observation space but is not practically possible due to the game's mechanics. The rest of the backdoor implementation was identical to that of Kiourti et al.~\cite{kiourti2020trojdrl} for strong targeted attacks (also employed by Bharti et al.~\cite{bharti2022provable}), which involves placing the backdoor at uniform intervals in the environment and rewarding the agent for behaving unsafely in its presence. 

\begin{figure}
    \centering
    \includegraphics[width=1\linewidth]{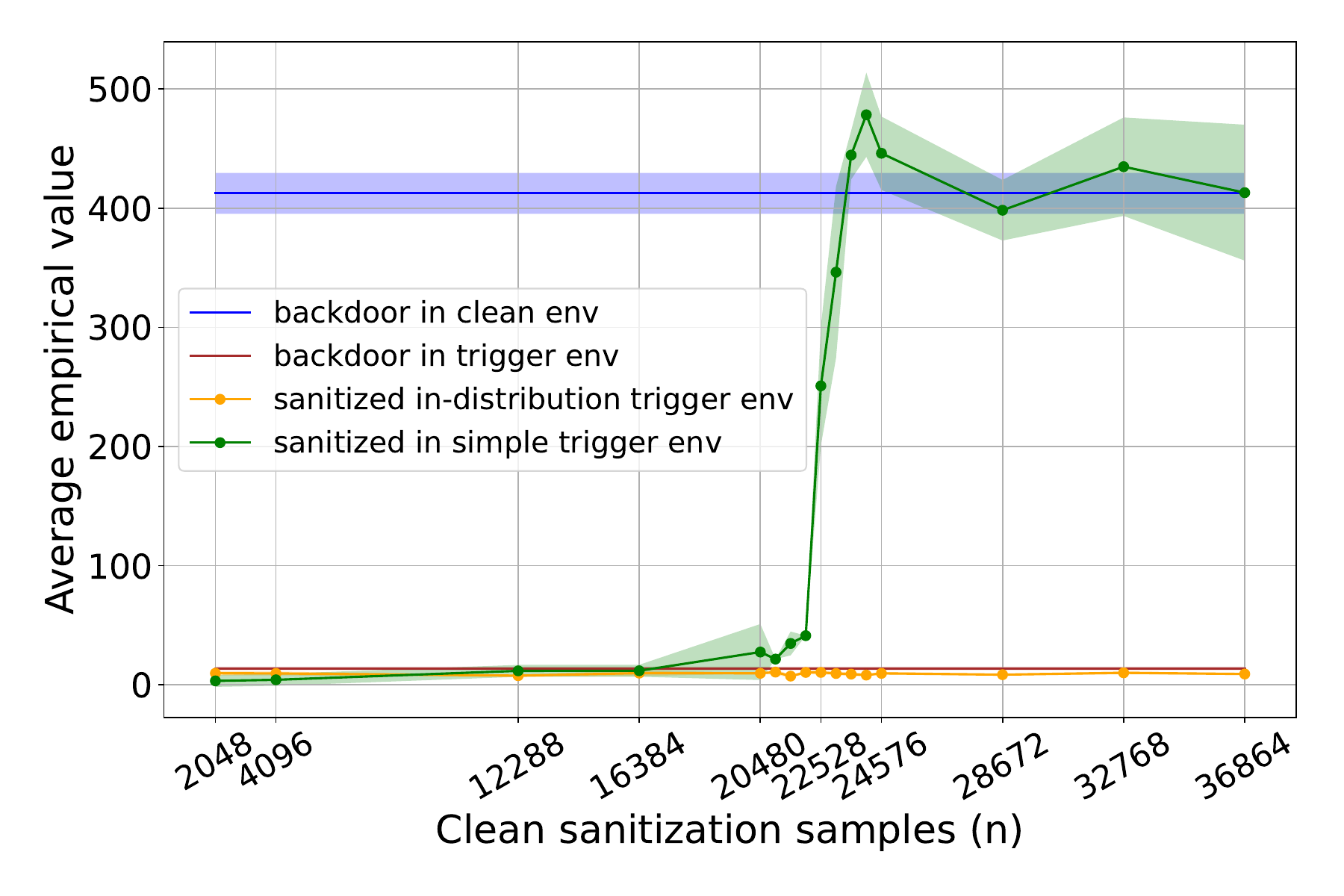}
    \caption{{Graph comparing the effectiveness of Bharti et al.'s~\cite{bharti2022provable} sanitisation algorithm against sample size, with agent performance baselines in clean (blue line) and simple trigger scenarios (red line). The algorithm's effect on neutralising a simple trigger is shown by the green line, while its impact on our in-distribution trigger is illustrated by the orange line. The results show that our in-distribution trigger eludes neutralisation by their algorithm, highlighting its inability to detect subtle triggers.\vspace{-2mm}}}
    \label{fig:performance_breakout_bharti}
\end{figure}

\begin{figure}
    \centering
    \includegraphics[width=1\linewidth]{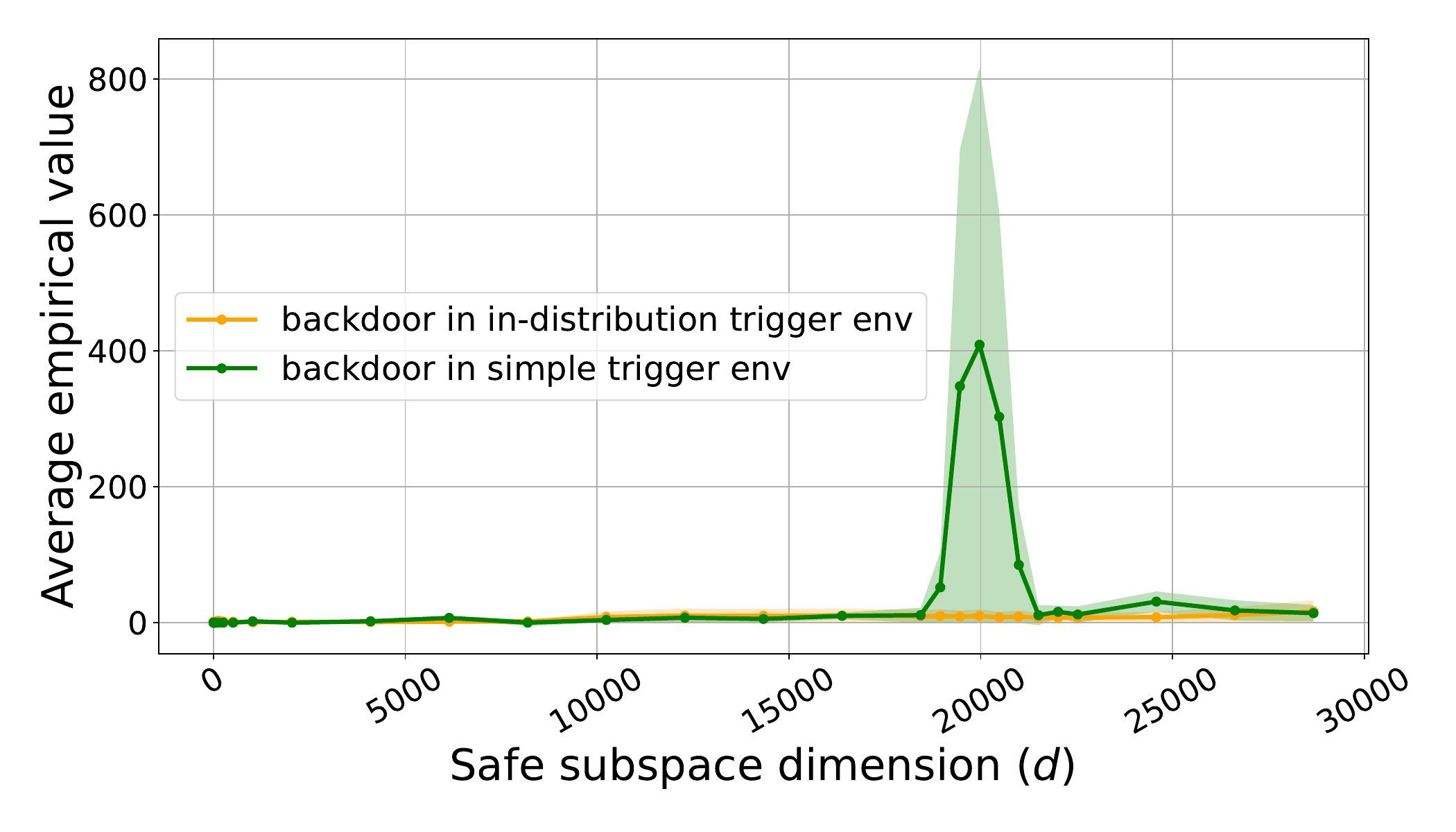}
    \caption{The graph shows the impact of Bharti et al.'s~\cite{bharti2022provable} sanitisation algorithm on agent behaviour with increasing empirical safe subspace dimensions across 32,768 samples. The green line shows how the algorithm retains the performance of the agent when the safe subspace has 20,000 dimensions, while simultaneously neutralising a simple backdoor trigger. The orange line depicts its performance when (unsuccessfully) attempts to neutralise our in-distribution trigger. This highlights that the in-distribution trigger is within the algorithm's safe subspace and evades the defence.\vspace{-4mm}}
    \label{fig:performance_breakout_bharti_dimensions}
\end{figure}

As depicted in Figure \ref{fig:performance_breakout_bharti}, our in-distribution backdoor successfully eludes the sanitisation algorithm of ~\cite{bharti2022provable}. Specifically, the authors' results against a simple trigger, represented in green, show the algorithm's performance surpassing that of the clean environment following sanitisation. However, when the environment incorporates our in-distribution trigger, there is no observed increase in the DRL agent's performance (indicated in orange) after undergoing the sanitisation algorithm's operation. This means that the
trigger was projected in the safe subspace and the backdoored agent was able to perceive it. Thus, the defence failed.

Moreover, we evaluate the impact of varying the number of dimensions related to the empirical estimation of the safe subspace, as implemented by the authors, depicted in Figure \ref{fig:performance_breakout_bharti_dimensions}. The results clearly show that changes in the dimensions do not substantially influence the agent's performance following the sanitisation phase. Although there is a minor improvement in performance after reaching 25,000 dimensions, there is no definitive proof suggesting further enhancements in performance for estimated dimensions beyond 28,000, which is the maximum dimension tested by the authors. Therefore, Figures \ref{fig:performance_breakout_bharti} and \ref{fig:performance_breakout_bharti_dimensions} validate our initial hypothesis that in-distribution triggers which lie within the empirical safe subspace cannot be effectively neutralised with the proposed method.

We observed that the sanitisation algorithm's execution time spanned several days, surpassing the agent's training duration. This indicates the approach's unsuitability for both rapid and slow-response scenarios in combating complex backdoor attacks. The necessity for immediate, effective, and ideally real-time countermeasures against backdoors in dynamic environments highlights the urgent need for more agile and robust solutions. In the next section, we propose a novel research direction aimed at offering a reliable alternative for mitigating elusive policy backdoors.

\section{Detection via Neural Activation Space\vspace{-2mm}}\label{sec:neural_act}
In our Threat Model (Section~\ref{sec: threat-model}), we discuss how victims might not recognize backdoors in targeted attacks but can observe DRL dynamics through neural activations. Drawing on \cite{liu2018trojaning, wang2019neural}, which identified distinct neural activations from backdoor triggers in supervised learning, we expand these insights to DRL. In contrast, our study examines the variations in neural activation patterns, influenced by the reward function, between episodic end goals and rare, concealed backdoors respectively. The unique neural activations in the presence of in-distribution triggers will enable us to identify the neurons linked to malicious actions, potentially unveiling how specific neurons activations temporally vary to execute these actions. \vspace{-2mm}
\newline
\subsection{Experimental Setup\vspace{-2mm}}
\begin{figure}[]
    \includegraphics[width=.24\textwidth]{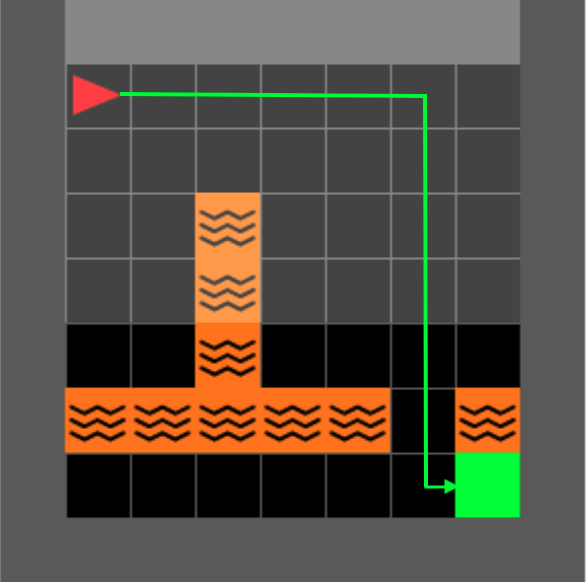}\hfill
    \includegraphics[width=.24\textwidth]{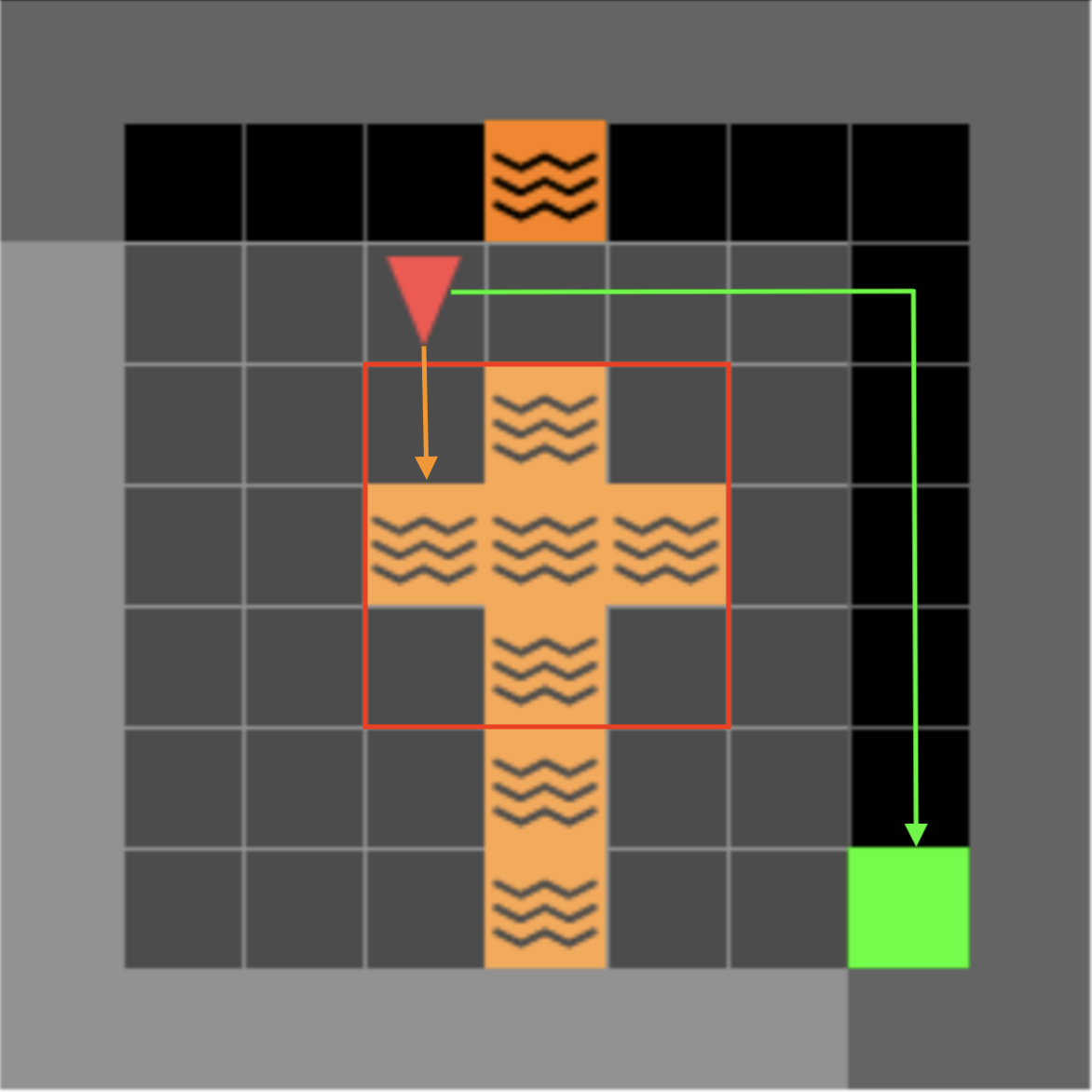}
    \caption{The visualisations illustrate our in-distribution trigger in the MiniGrid Crossings environment. From left to right, the images show: a) the environment without a trigger, and b) the environment with a ``+''shaped trigger (red box). In (a), the backdoored agent reaches the goal safely, whereas in (b), it walks into a lava block as the trigger is present.\vspace{-4mm}}
    \label{fig:visualised minigrid environment}
\end{figure} 

To test if backdoor triggers can be identified in neural activation space, we employ the Parameterised LavaWorld environment by Ashcraft et al.\cite{ashcraft2021poisoning}, an adaptation of the MiniGrid-LavaCrossingsS9N1-v0 from gym-minigrid\cite{MinigridMiniworld23}. This environment, chosen for its high variability from randomly placed lava 'rivers' and a square-based design conducive to hiding backdoor triggers, requires the agent to navigate from start to finish, avoiding a randomly positioned wall of lava. Unlike MiniGrid-LavaCrossingsS9N1-v0, it features an extra row of three lava squares, adding complexity. The main lava 'river,' safe passage, and additional lava squares vary with each episode, creating a dynamic challenge for the agent whose receptive field extends 7 squares forward and 3 squares to both sides (Figure \ref{fig:visualised minigrid environment}a).

Following the sanitisation method outlined in Section~\ref{sec:sanitisation}, we trained two agents using PPO and a Convolutional Neural Network: one benign and one backdoored, over 60 million frames in 10 parallel environments. The backdoor activates when extra lava squares align with the main lava 'river' to form a cross, a setup inspired by previous work~\cite{kiourti2020trojdrl,ashcraft2021poisoning, bharti2022provable}, with the trigger mimicking the real goal's reward but in a short-term context (Figure~\ref{fig:visualised minigrid environment}b). Post-training evaluation on 1,000 trigger-free episodes showed approximately 95\% accuracy for both agents. Analysis of the Actor network's 256 neurons during 1,000 goal-visible episodes (3682 samples) and 1,000 trigger episodes (3219 samples) then followed.

\subsection{Results\vspace{-2mm}}
\begin{figure}
    \centering
    \includegraphics[width=0.85\linewidth]{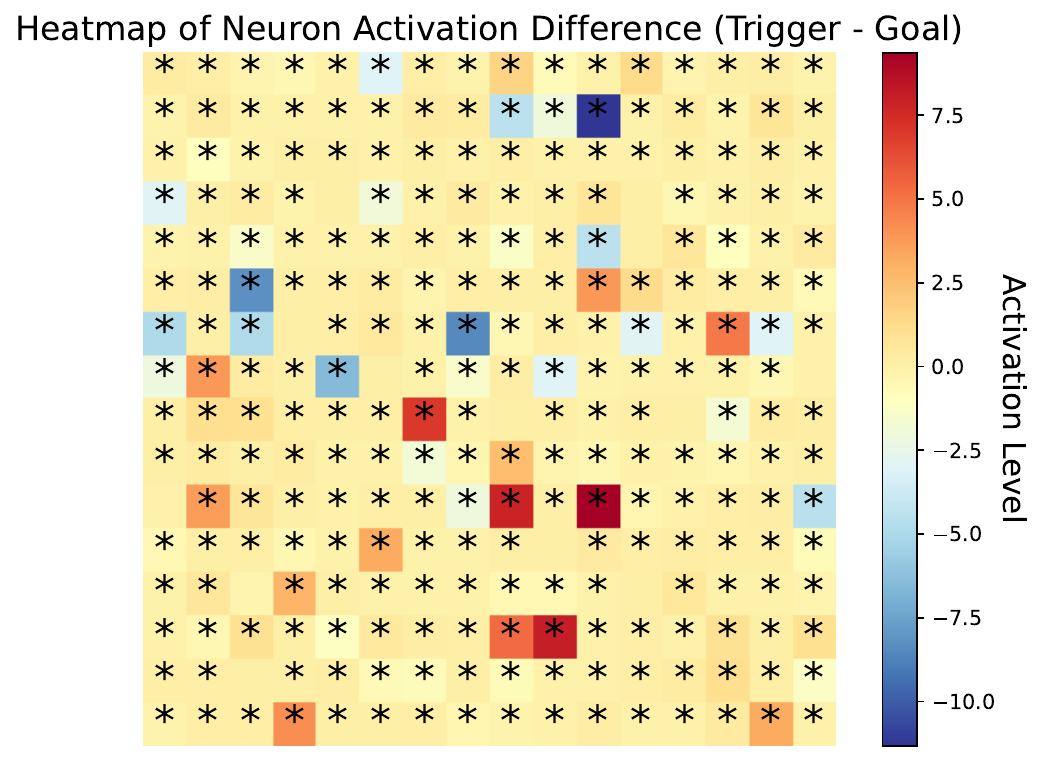}
    \caption{The 16x16 heatmap shows variations in PPO's actor network neuron activations between two scenarios: 1) with an in-distribution trigger visible, and 2) with the goal visible. Darker red signals indicate a stronger neuron response to the trigger, whereas darker blue signify a stronger neuron response to the goal. This efficiently demonstrates the fluctuation in neuron activations due to in-distribution triggers. ("*" denotes statistical significance.)\vspace{-4mm}}
    \label{fig:difference in activations trigger goal}
\end{figure}

Figure~\ref{fig:difference in activations trigger goal} illustrates the notable variance in neural activations between episodes with the goal and with the trigger. A significant disparity is observed, particularly in specific neurons. The Mann-Whitney U-test confirms the statistical significance of these differences (refer to Figures ~\ref{fig:heatmap_average_difference_goal_trigger} and \ref{fig:histogram-distribution-statistically-significant} in the Appendix). This evidence suggests potential for distinguishing benign from malicious activations, guiding the development of a trigger detection mechanism. Further insights can be found in the Appendix (Figures~\ref{fig:histogram-distribution-non-statistically-significant} and 12).

\section{Trigger Classifier\vspace{-2mm}}\label{sec:classifier}
Based on our findings, we now design a simple classifier to detect the presence of triggers. As explained in Section ~\ref{sec: threat-model}, the defender does not have access to any information about the triggers. Hence, we need our classifier trained only on episodes from the clean environment. We set up nine straightforward classifiers based on the activation levels of 64 neurons from 10,000 episodes without triggers. These classifiers use a mix of high (99.5, 99, and 98 percentiles) and low (0.5, 1, and 2 percentiles) quantiles from these activation levels as a reference point to spot triggers. Additionally, we apply 80 different threshold levels (from 1 to 80) across these classifiers to decide whether an episode was normal or had been tampered with by a trigger. For instance, if the number of neuron activations that fell outside our set quantile thresholds exceeded a certain threshold level, the detector would classify that episode as abnormal (i.e., a trigger is present). A threshold of 80 for a network of 256 neurons was selected to understand the trade-off between true positives and false positives. 

\begin{figure}
    \centering
    \includegraphics[width=1\linewidth]{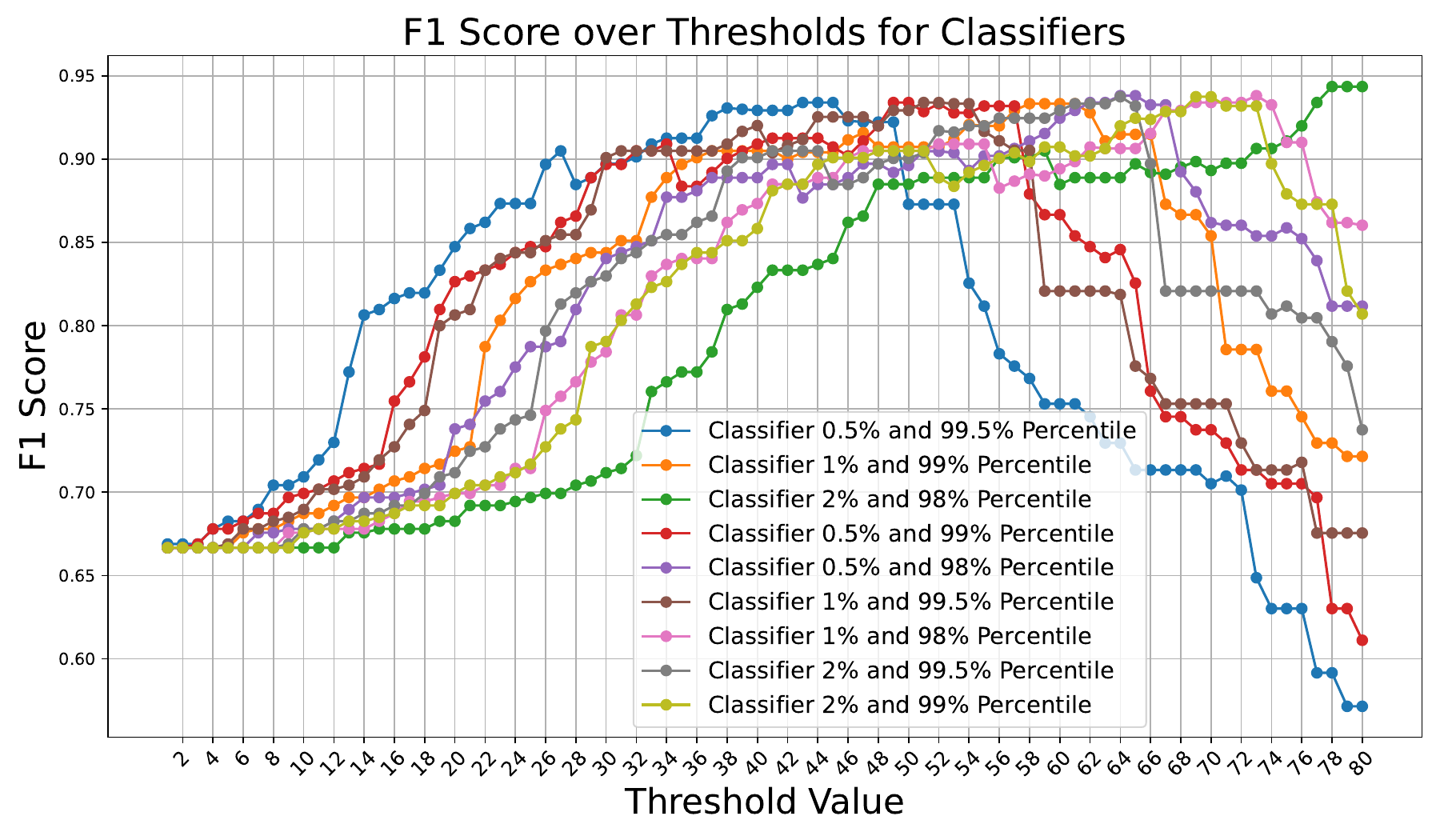}
    \caption{The figure presents the F1 Scores for all 9 simple classifiers across various thresholds, demonstrating the potential to detect backdoors via neuron activations by leveraging the distribution of activation levels in the PPO actor network. This indicates that analysing neuron activation patterns can be an effective method for identifying backdoors.}
    \label{fig:f1-scores}
\end{figure}
\begin{figure}
    \centering
    \includegraphics[width=1\linewidth]{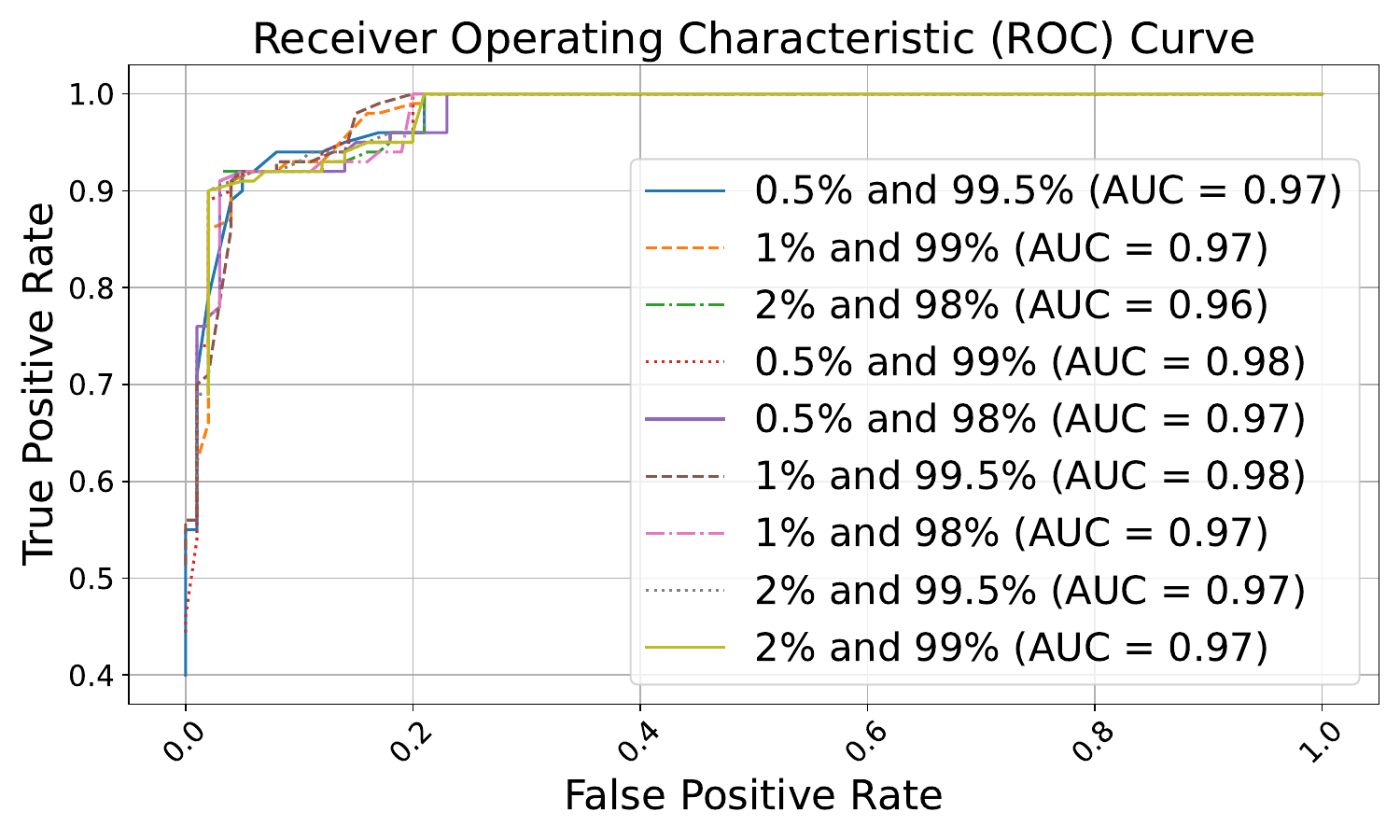}
    \caption{The figure presents the ROC Curves for all 9 simple classifiers across various thresholds. The use of thresholds allow us to assess the true positive and false positive rates for all simple detectors. The detectors showed AUC values as high as 0.98 without the consideration of episodic temporality in the detectors. \vspace{-1mm}}
    \label{fig:ROC-curves}
\end{figure}

As illustrated in Figure \ref{fig:f1-scores}, the most effective detector, specifically those configured with thresholds at the 2/98\% quantiles, achieved an F1 score of 0.94. This indicates a commendable balance between precision and recall, highlighting its efficiency in identifying true positives without excessively misclassifying negatives. Remarkably, the classifier with thresholds set at 2\%/98\% quantiles demonstrated a true positive rate of 92\% and false positive rate of 3\% reinforcing its effectiveness. This is further proven in the Appendix's Figure ~\ref{fig:histogram-distribution-statistically-significant} and ~\ref{fig:histogram-distribution-non-statistically-significant} which shows the distribution of neuron activations in the presence of triggers and goals respectively.  

Figure \ref{fig:ROC-curves} reveals that the majority of our thresholding classifiers exhibit an Area Under the Curve (AUC) greater than 0.95, with the highest recorded AUC value at 0.98. This performance metric underscores the capability of these classifiers to reliably detect in-distribution backdoor triggers across a significant majority of episodes in which such triggers are present. Notably, this detection efficacy is achieved using a very inexpensive classifier (with minimal computational overhead) and could be further improved using more advanced algorithms depending on the computational capacity available. 

\section{Related Work}\label{sec: rw}
Liu et al~\cite{liu2018trojaning} and Shafahi et al~\cite{shafahi2018poison} demonstrate LSTM backdoors that redirect agents upon trigger activation, reducing performance. Kiourti et al~\cite{kiourti2020trojdrl} present TrojDRL, showing DRL's vulnerability to backdoors without compromising clean task performance by altering observations via a man-in-the-middle attack, leading to changed agent behaviour upon trigger. Gunn et al~\cite{gunn2022adversarial} and Yu et al~\cite{yu2022temporal} explore RL poisoning in application-specific scenarios, showing adversarial perturbations during training slow agent. The latter achieves this by using temporal DRL attacks 
Wang et al~\cite{wang2021backdoorl} present multi-agent RL backdoor attacks, significantly decreasing a victim's win rate from 37\% to 17\% through competitor-triggered backdoors. Chen et al~\cite{chen2022marnet} and Foley et al~\cite{foley2022execute} propose a DRL poisoning strategy that causes misbehaviour in specific states by altering some training observations, the latter publication proven effective in Atari game tests. Rakhsha et al~\cite{rakhsha2020policy} create an optimisation framework for stealthy attacks in RL, showing target policy imposition on victims. Ashcraft et al~\cite{ashcraft2021poisoning} develops an in-distribution DRL backdoor models, corrupting policies during inference. The TrojAI challenge by US IARPA and NIST (\url{https://pages.nist.gov/trojai/docs/index.html}) aims to advance DRL backdoor defence, using DRL agents on MiniGrid-LavaCrossingS9N1-v0 with in-distribution triggers and 300 models for training and testing. Participants develop backdoor detectors, but the intensive training demand restricts its practicality in critical sectors like autonomous driving, where numerous model iterations are beyond the affordability of end-user resources.

Despite the criticality of the problem, limited works have proposed solutions for detecting or mitigating backdoors for DRL agents. As discussed in Section~\ref{sec:sanitisation}, Bharti et al.\cite{bharti2022provable} presented a defence algorithm based on a wrapper method around the backdoor policy that provides performance guarantees against all subspace trigger based adversaries. However, as we demonstrate in Section~\ref{sec:sanitisation} their solution does not generalise to adversaries using in-distribution triggers. Acharya et al.~\cite{acharya2023universal} introduced an attribution analysis-based algorithm to detect DRL backdoors, exploiting advantage prediction sensitivities to observation changes. They used Jacobian matrices to identify trigger-affected inputs, showing success in IARPA's TrojAI (\textit{rl-lavaworld-july2023} and \textit{randomised-lavaworld-aug2023}). However, the approach lacks practicality due to the extensive training requirement of the challenge. Guo et al.~\cite{guo2023policycleanse} introduced PolicyCleanse for backdoor detection and mitigation in Competitive RL (CRL), using reward reversal to identify and counteract opponent-triggered backdoors. It generates a Trojan policy to mimic potential triggers, reverses the reward function, and evaluates the target for malicious behaviour. Triggers are mitigated by training the victim with benign and pseudo-trigger episodes. Despite its effectiveness, PolicyCleanse's high computational needs and specificity to CRL environments constrain its wider applicability, especially for real-time policy detection and scenarios involving human interaction.

\section{Conclusions \& Future Work}
In this work, we evaluate the effectiveness of existing DRL backdoor mitigation strategies, revealing their limited generalizability against sophisticated threats. Our investigation into neural activation spaces for identifying harmful triggers introduces a precise, efficient classifier, marking a novel path in backdoor detection research. This opens avenues for future work to extend these insights across different algorithms and settings, and to examine classifiers that assess neural activation patterns temporally.

\newpage
\section{Acknowledgements}
Research funded by the Defence Science and Technology Laboratory (DSTL) which is an executive agency of the UK Ministry of Defence providing world class expertise and delivering cutting-edge
science and technology for the benefit of the nation and allies. The  research supports the Autonomous Resilient Cyber Defence (ARCD) project within the Dstl Cyber Defence Enhancement programme.

\bibliography{references}
\bibliographystyle{IEEEtran}


\onecolumn

\appendix
\subsection{Statistical Analysis of Activation Patterns}

\begin{figure}[hbt!]
    \centering
    \includegraphics[width=.4\textwidth]{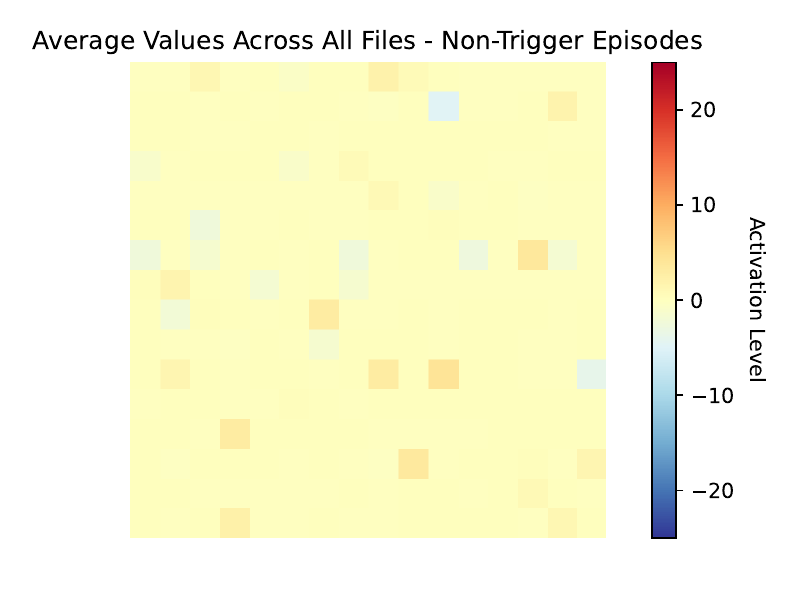}\hfill
    \includegraphics[width=.4\textwidth]{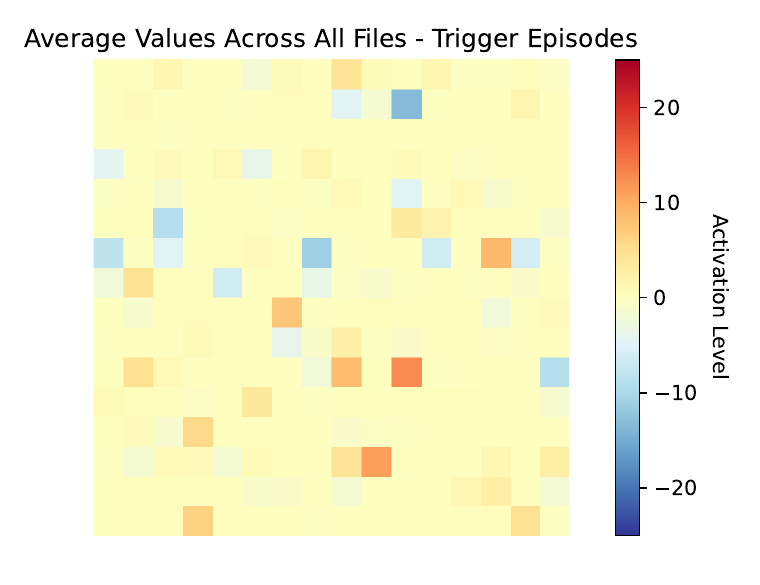}\hfill

\caption{Heatmap of average neuron activations a)  in a triggered environment b) in a non-triggered environment. The figure allows us to differentiate between triggered and non-triggered environments through the overall episodic neural activations. This allows us to create a base case that signifies how the presence of a trigger affects the agent through the neural activation space. }
\label{fig:heatmap_triggered_non_triggered_environment}
\end{figure}

\noindent The data depicted in the figures provides an insight into the dynamics of neural activation within the actor network of Deep Reinforcement Learning (DRL) agents, particularly under varying environmental conditions. Figure ~\ref{fig:heatmap_triggered_non_triggered_environment} offers a visual comparison between the neural activation patterns in triggered versus non-triggered environments. This distinction is crucial, as it highlights the impact that in-distribution triggers have on the neural activation space of the agent. By examining the intensity and distribution of neural activations, we observe how certain neurons become more active (positively and negatively) in the presence of environmental triggers compared to standard conditions.

\begin{figure}[]
    \centering
    \includegraphics[width=.4\textwidth]{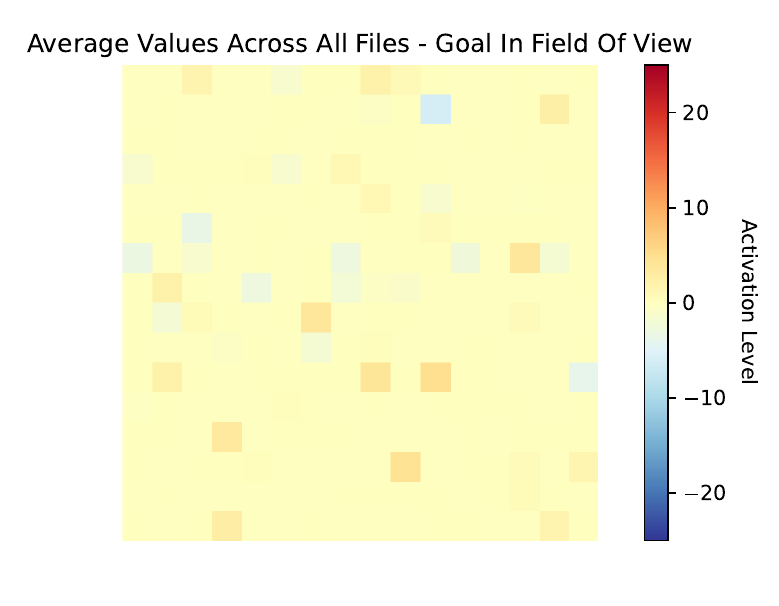}\hfill
    \includegraphics[width=.4\textwidth]{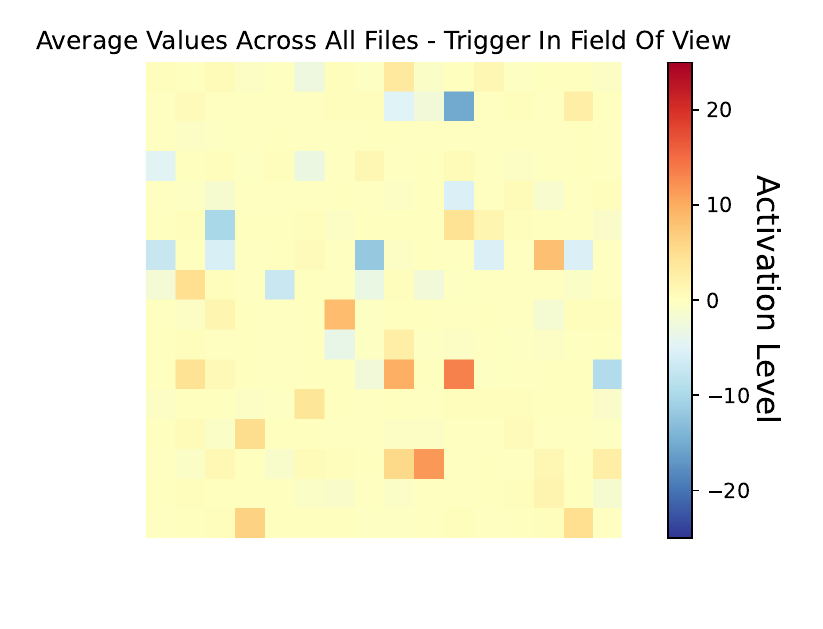}\hfill

\caption{Heatmaps of average neuron activations when a) when the goal is in the field of view (in a non-triggered environment) and b) The trigger is in the field of view (in a triggered environment). Specific neuron activation values are greater during the presence of a trigger as compared to the presence of a goal, indicating the significance of the presence of triggers within the neural activation space.}
    \label{fig:heatmap_average_difference_goal_trigger}
\end{figure}

\noindent Furthering this observation, Figure ~\ref{fig:heatmap_average_difference_goal_trigger} delves deeper by contrasting the neural activation levels when a trigger is visible within the agent's field of view against when a goal object appears. The heatmaps derived from these scenarios showing the differences in neuron activation, where the presence of a trigger suggests a stronger neural response compared to the goal in the field of view. This differential activation pattern shows the influence of in-distribution triggers on the DRL agent's neural architecture, providing a foundation for understanding how triggers overall can manipulate agent behaviour through the neural activation space.

\noindent Together, these visualisations not only allow for a detailed analysis of the agent's response to different environmental stimuli but also establish a baseline understanding of how triggers can distinctly alter neural activation patterns. This knowledge paves the way for further exploration into the mechanisms through which DRL agents can be influenced or manipulated, highlighting the importance of neural activation analysis in identifying and understanding the impact of in-distribution of triggers on agent behaviour.

\subsection{Distribution of Neuron Activations}
\begin{figure}[hbt!]
    \centering
        \includegraphics[width=.5\textwidth]{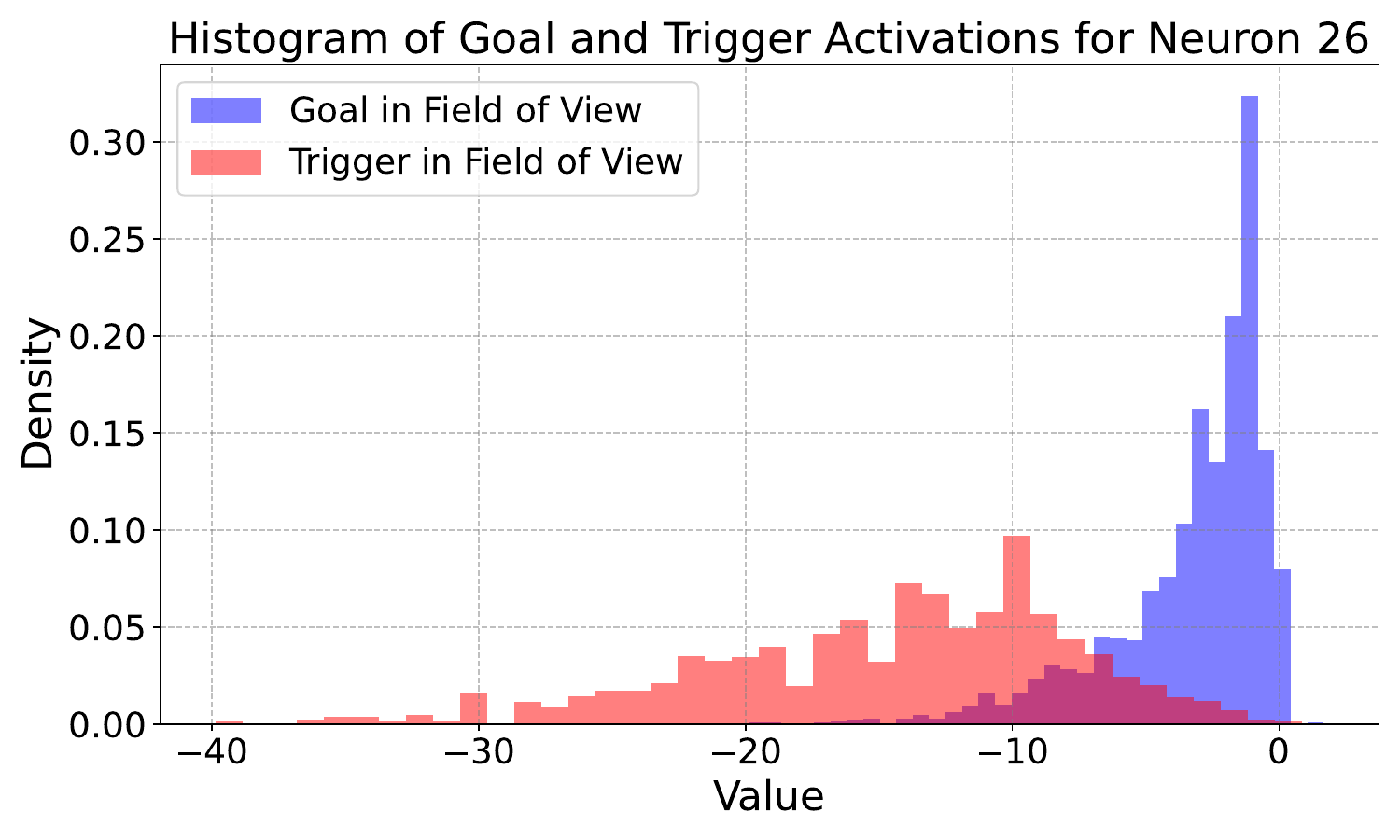}\hfill
            \includegraphics[width=.5\textwidth]{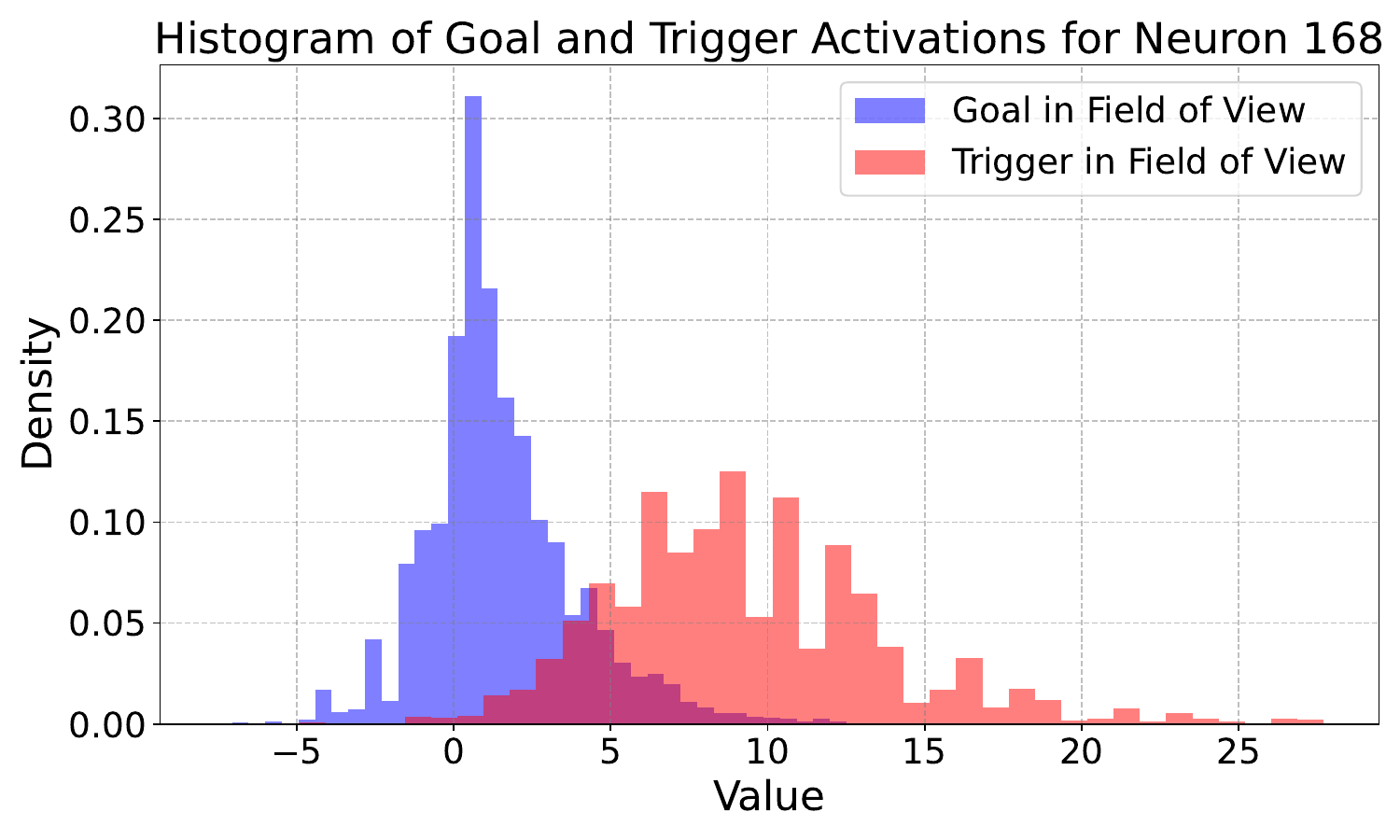}\hfill
        \includegraphics[width=.5\textwidth]{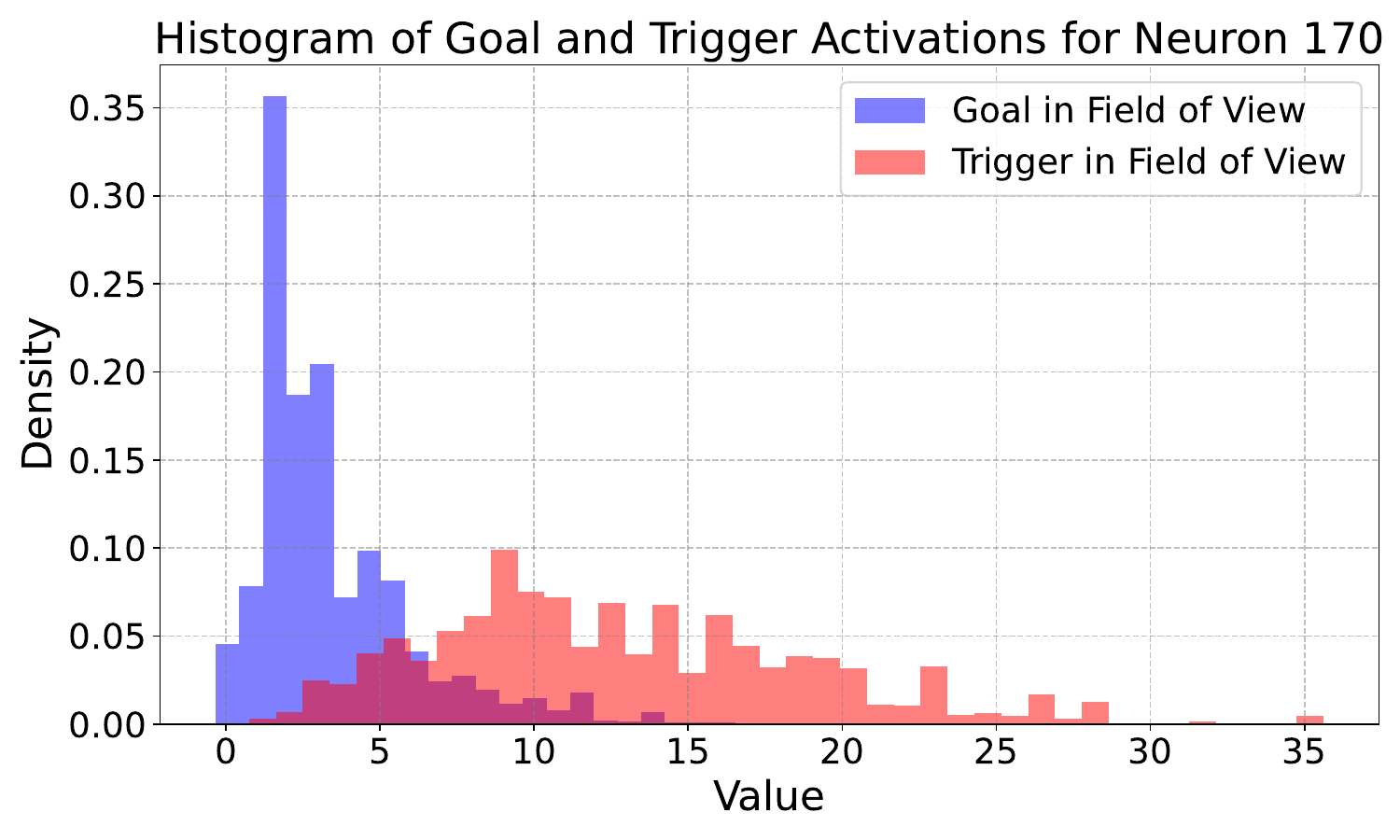}\hfill
    \includegraphics[width=.5\textwidth]{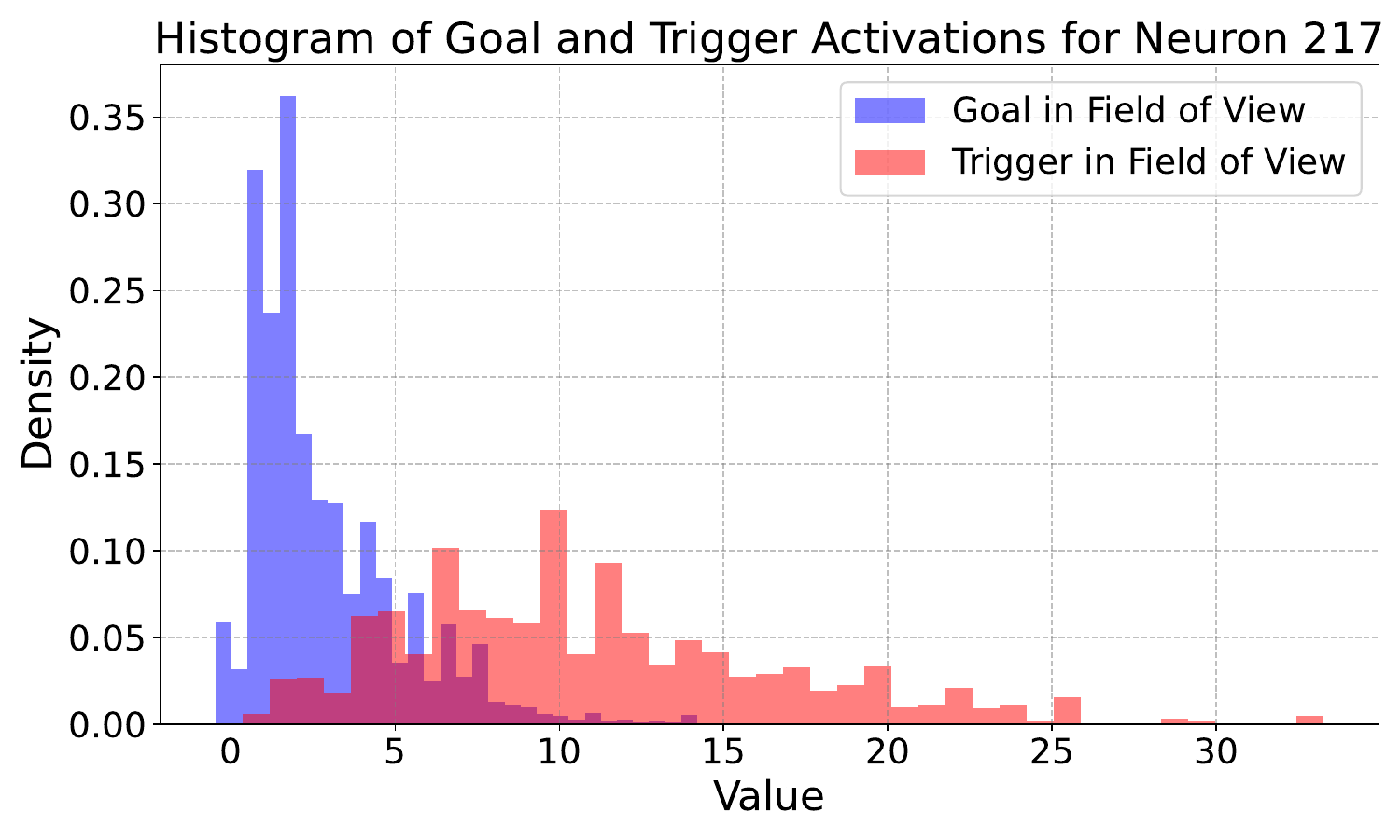}\hfill

\caption{Distribution of neuron activations levels for the most affected neurons within of PPO's actor network in MiniGrid when compared within scenarios including a) Goal in Field of View and b) Trigger in Field of View. The distribution suggests significant differences in the specific neural activations in the presence of a trigger in the field of view, as compared to when goal is in field of view. This is further backed up by the distributions of both being statistically significant to each other.}
    \label{fig:histogram-distribution-statistically-significant}
\end{figure}

\begin{figure}[]
    \centering
        \includegraphics[width=.5\textwidth]{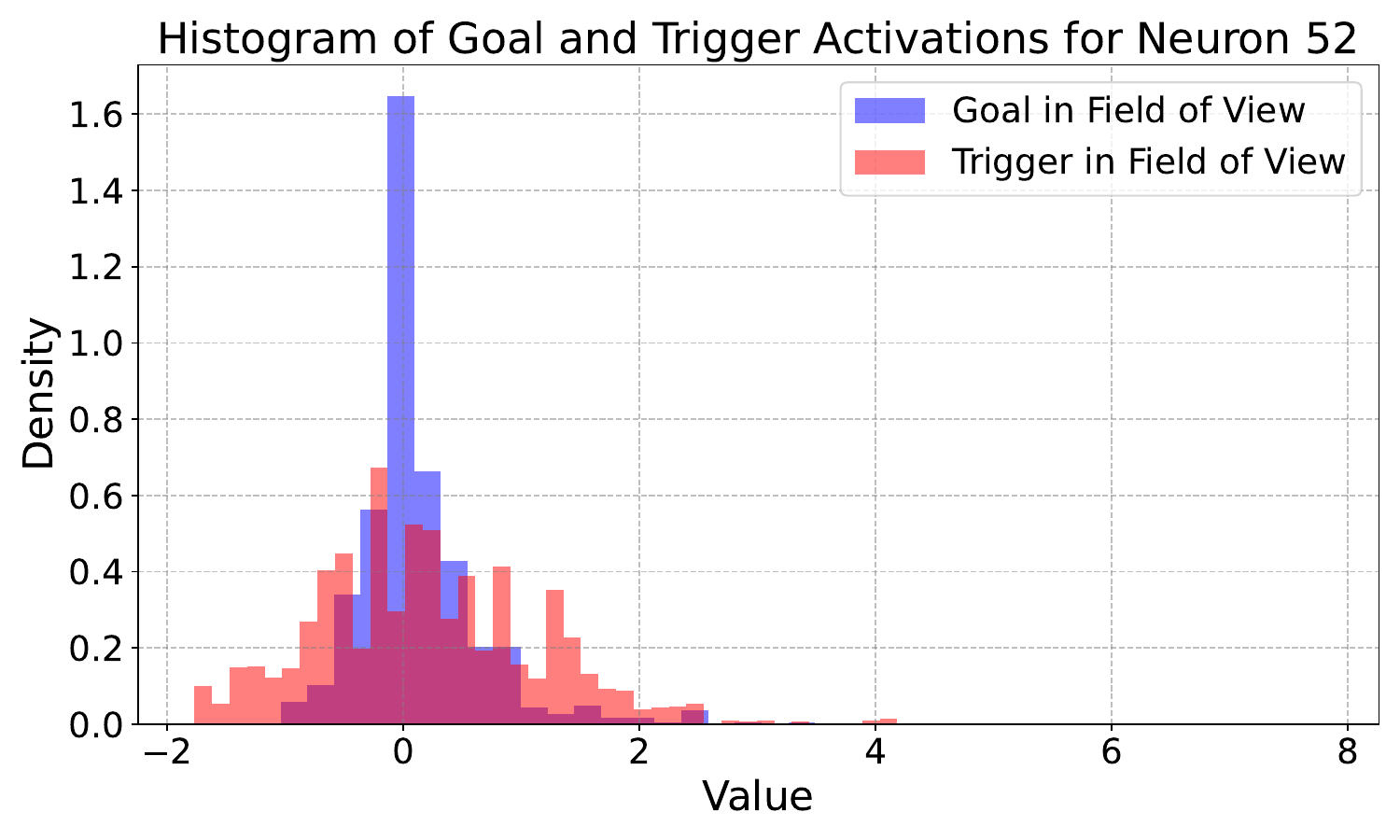}\hfill
            \includegraphics[width=.5\textwidth]{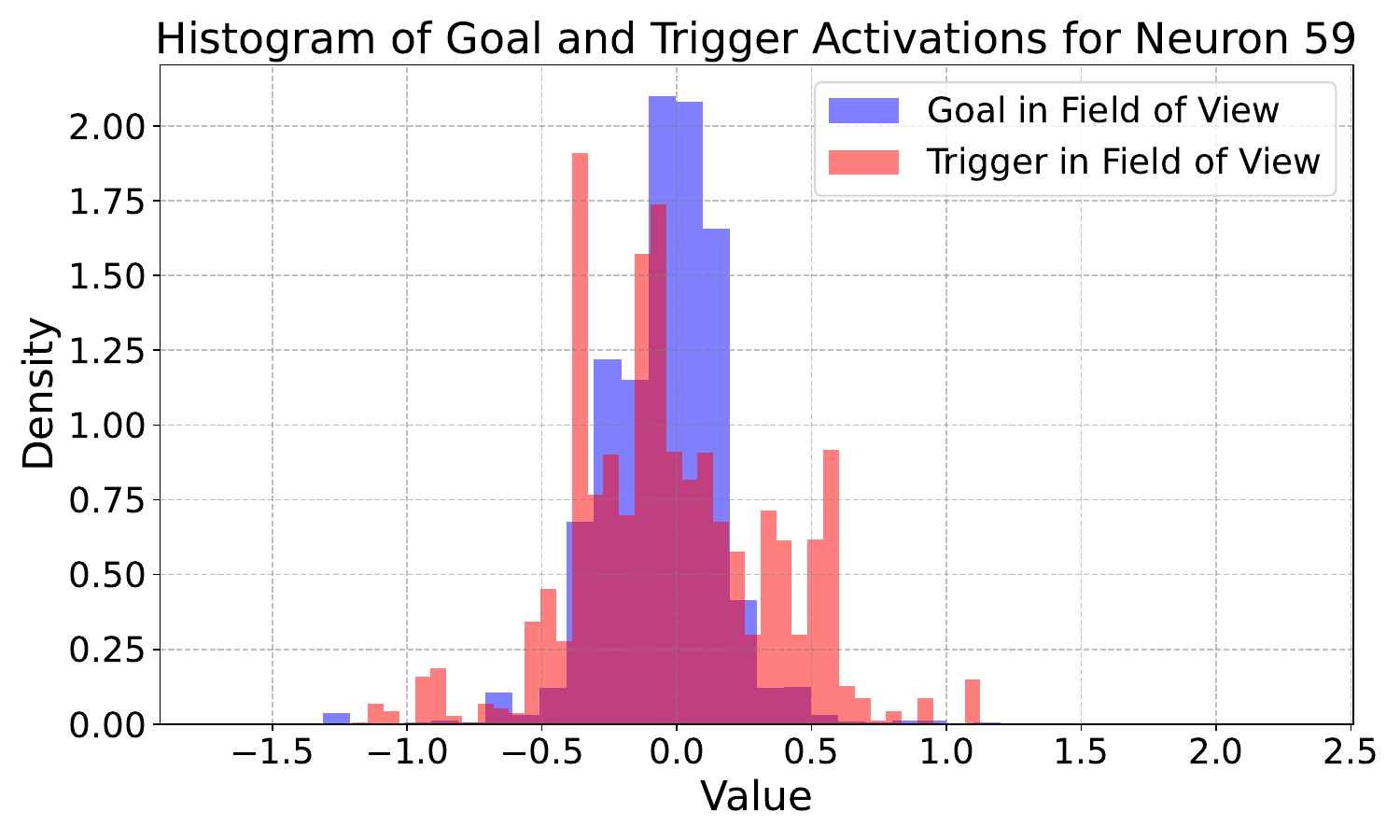}\hfill
        \includegraphics[width=.5\textwidth]{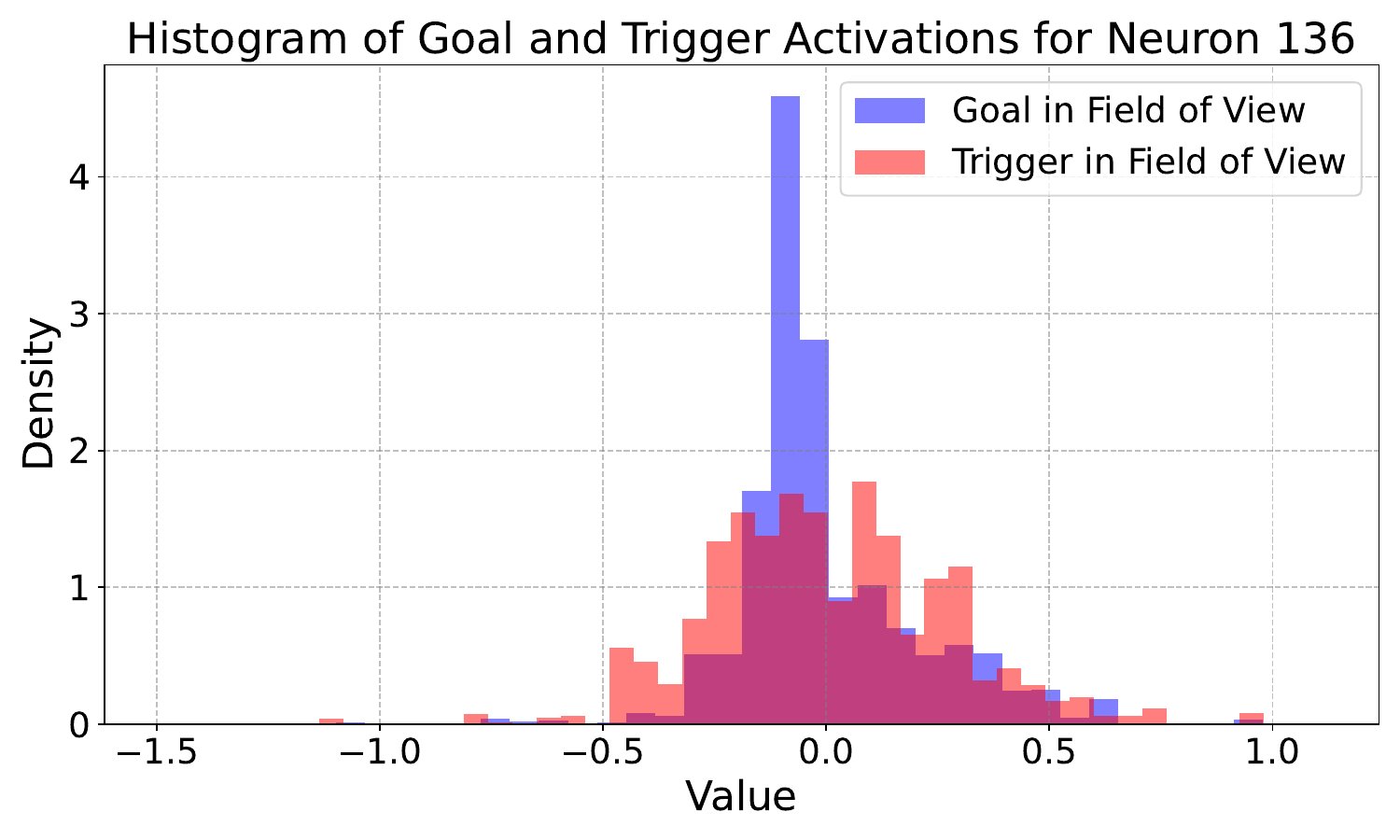}\hfill
    \includegraphics[width=.5\textwidth]{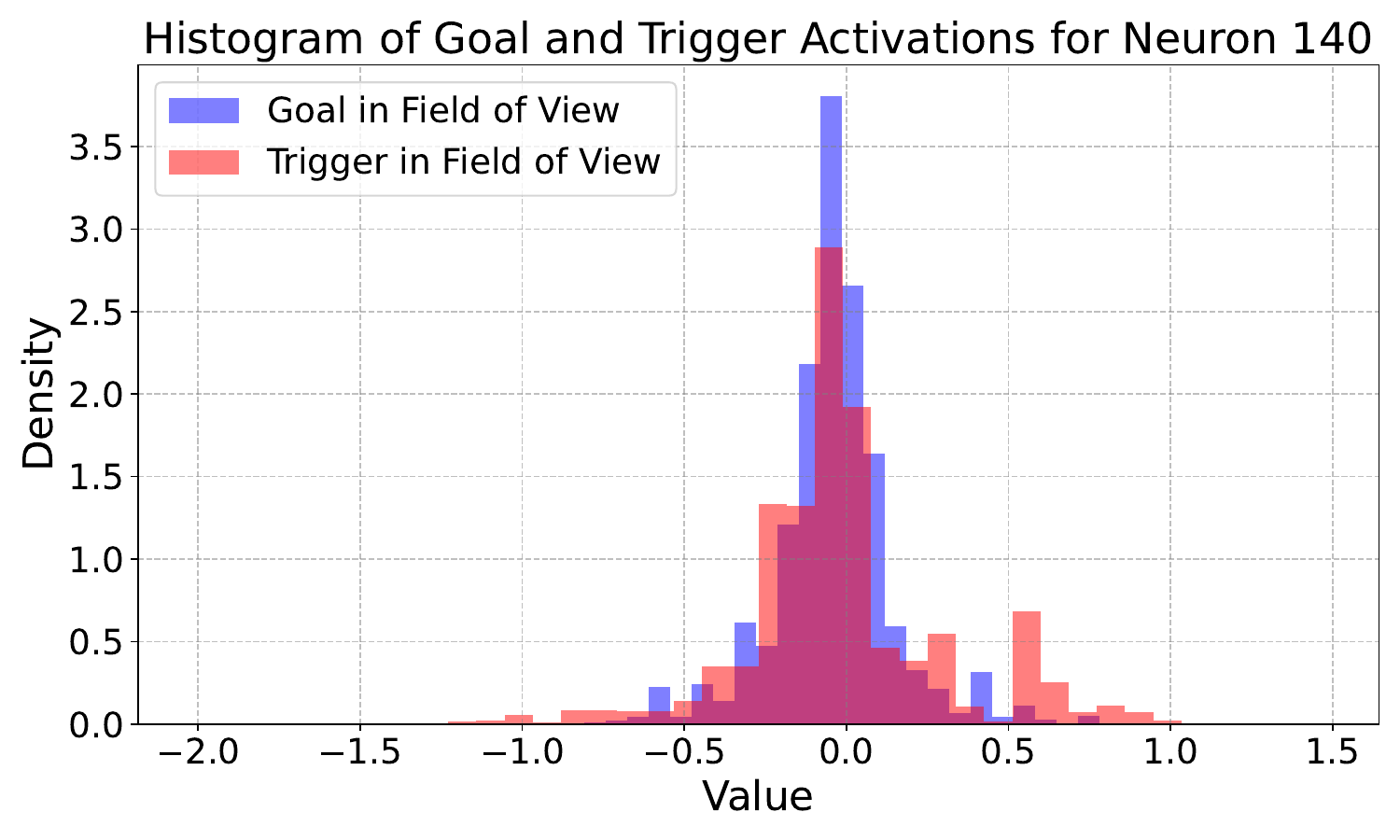}\hfill
        
\caption{Distribution of neuron activations levels for the least affected non-statistically significant neurons within of PPO's actor network in MiniGrid when compared within scenarios including a) Goal in Field of View and b) Trigger in Field of View. The distribution and medians of both scenarios suggest that the particular neurons are not affected by the presence of the trigger in the field of view, as compared to when goal is in field of view. This is further backed up by the Mann-Whitney U-test, which calculated the neurons activations to not be statistically significant}
    \label{fig:histogram-distribution-non-statistically-significant}
\end{figure}

\noindent Figure~\ref{fig:histogram-distribution-statistically-significant} delves deeper into the specific neuron activation distribution of neurons with higher activation levels shown in Figure ~\ref{fig:heatmap_average_difference_goal_trigger}. On the other hand, Figure~\ref{fig:histogram-distribution-non-statistically-significant} focuses on neurons that were not considered statistically significant within the two scenario's presented in Figure~\ref{fig:heatmap_average_difference_goal_trigger}. From Figure~\ref{fig:histogram-distribution-statistically-significant} It can be observed that specific neuron, when compared in the two scenarios, Trigger in field of view and Goal in field of view, show significant differences in their neuron activations distribution. Backing our hypothesis that the neuronal activation space can be utilised within DRL to detect elusive backdoors like in-distribution triggers.

\noindent Figure~\ref{fig:histogram-distribution-non-statistically-significant} explores the neuron activation levels of the neurons that were not statistically significant within the neuron activation space. It can be observed that the neuron activation distribution and medians of such neurons is similar in both, goal in field of view and trigger in field of view. This backs why Mann-Whitney U-test did not classify the selected neurons as statistically significant. 

\newpage
\noindent Figure 12 looks at the neuron activation level distribution of the neurons that did not show high levels of activation differences within the two scenarios. However, they still suggested statistical significance according to the Mann-Whitney U-test. As can be observed, it could be because the distribution levels and median slightly differ within the two scenarios. \\
\\
\noindent In all, it can be observed that the scenario, Trigger in field of view, shows a larger distribution of neuron activation levels when compared to the Goal in field of view scenario. This discovery could also act as a point of research to develop more complex detection methods against a variety of different backdoors.   

\begin{figure}
    \centering
        \includegraphics[width=.5\textwidth]{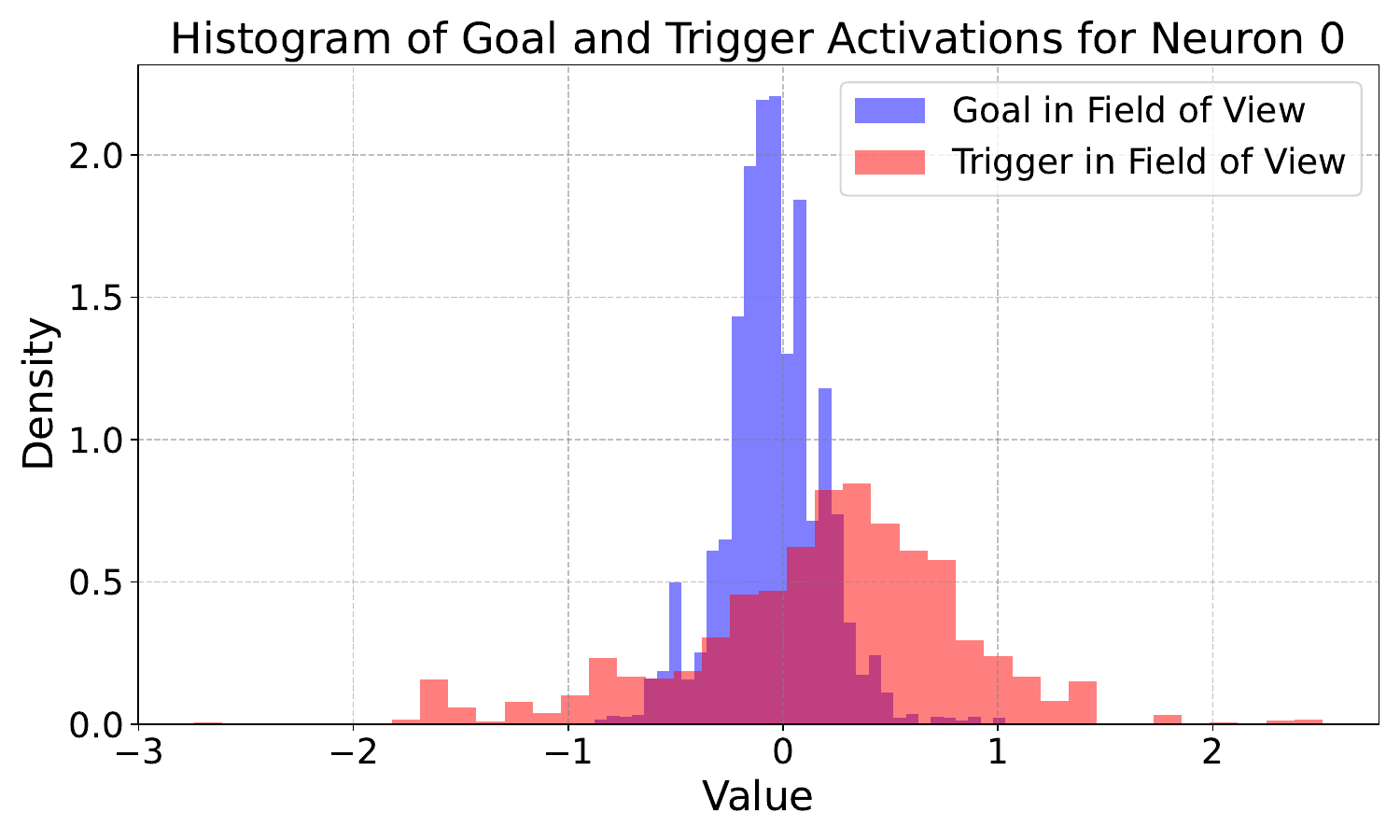}\hfill
            \includegraphics[width=.5\textwidth]{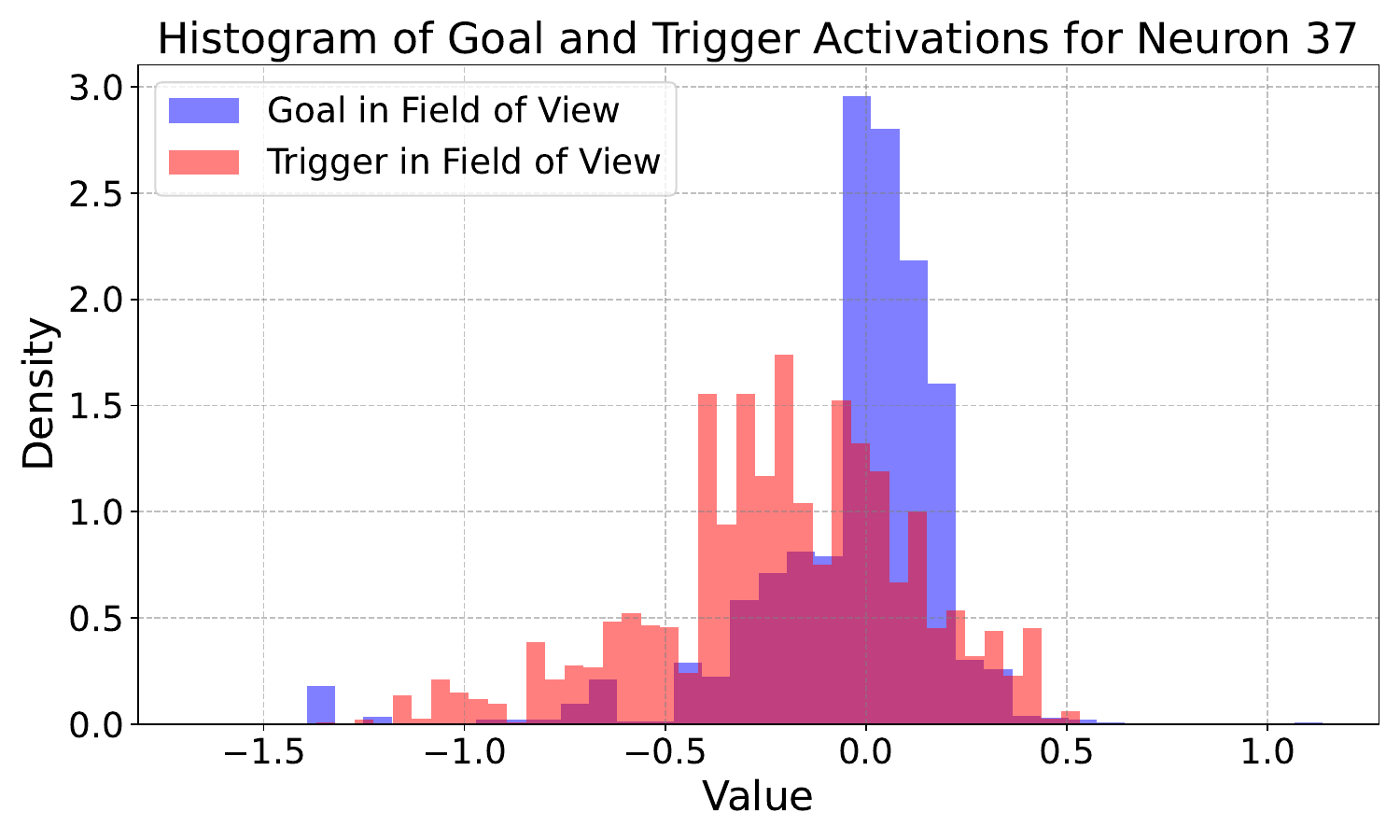}\hfill
        \includegraphics[width=.5\textwidth]{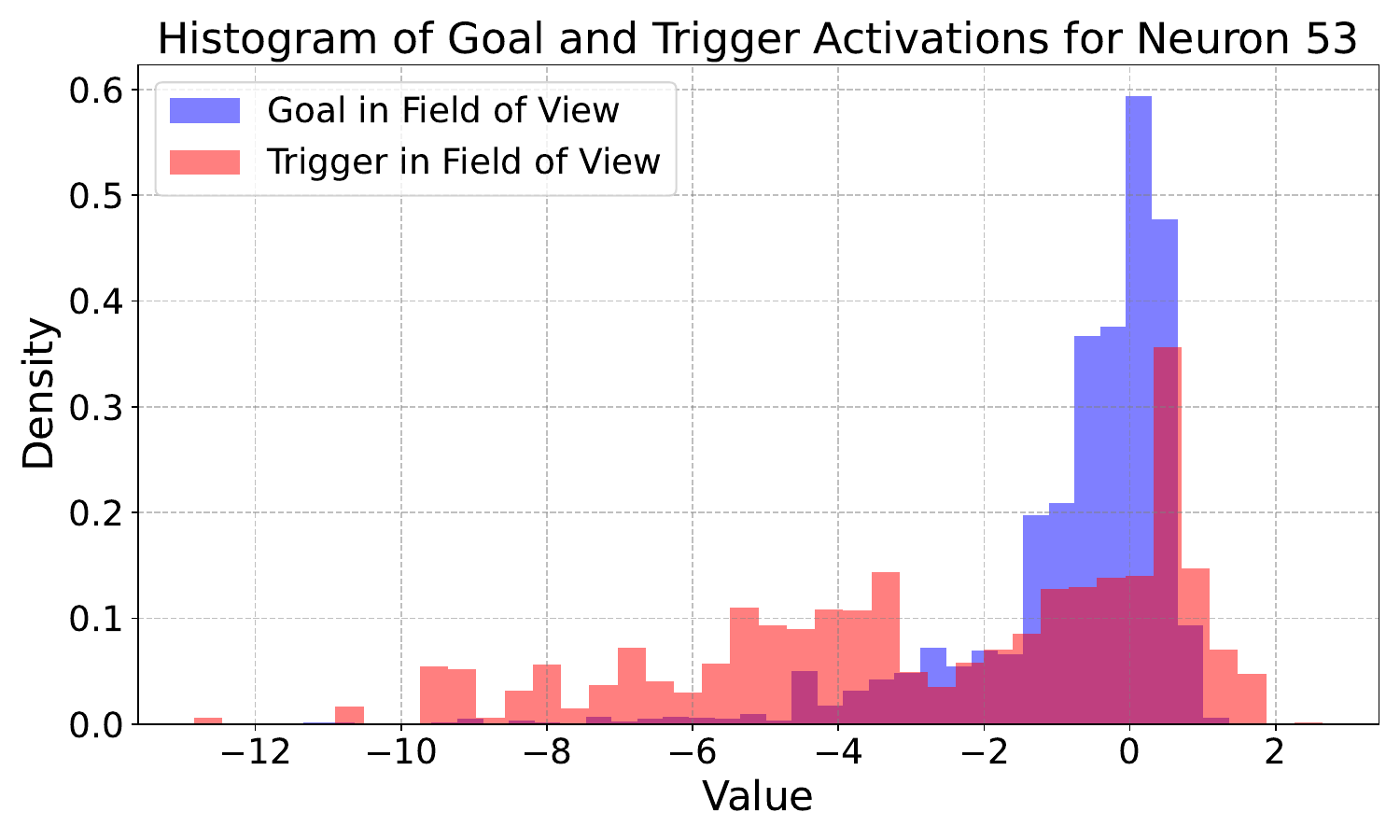}\hfill
    \includegraphics[width=.5\textwidth]{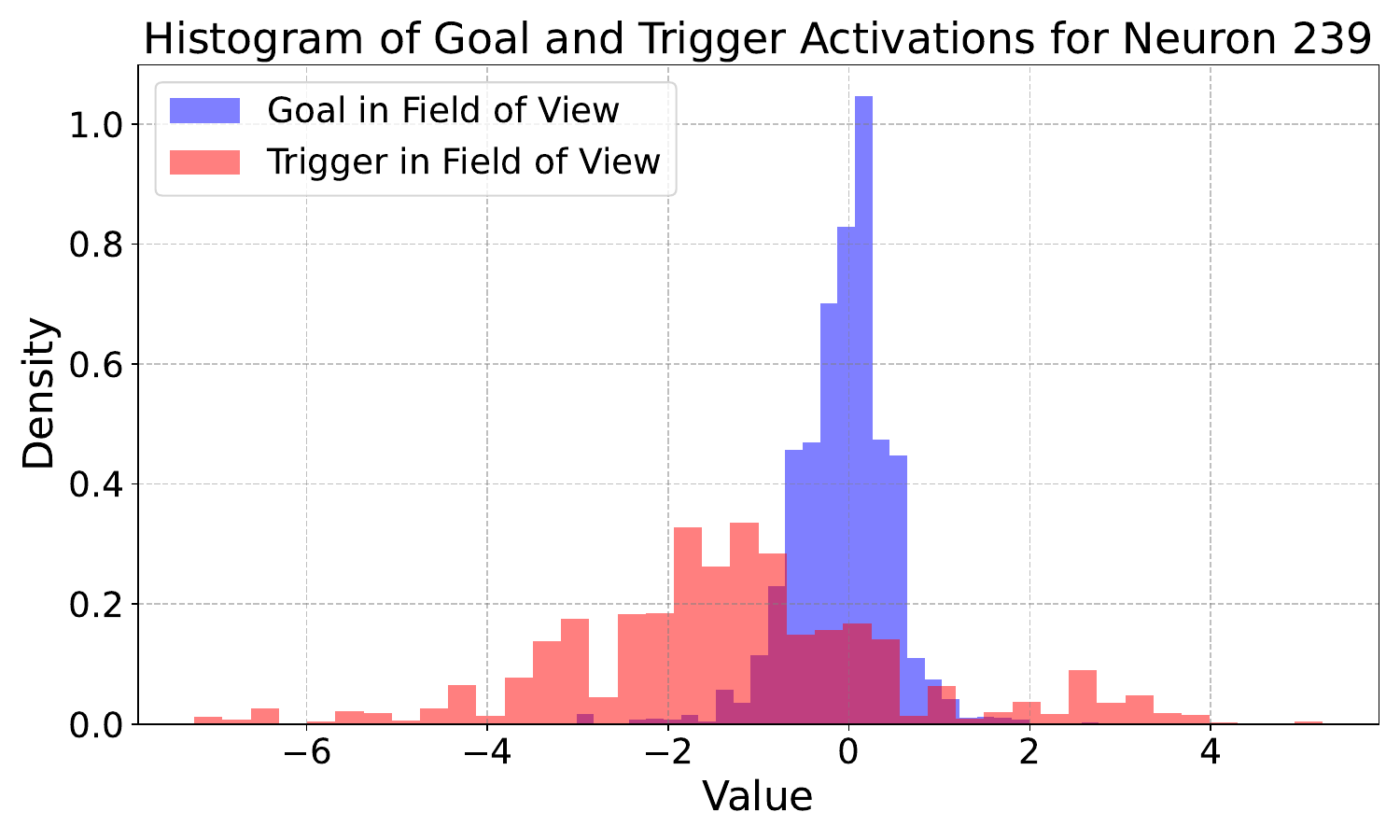}\hfill
\label{fig:histogram-distribution-statistically-significant-less-activated-neurons}
\caption{Distribution of neuron activations levels for the least affected statistically significant neurons within of PPO's actor network in MiniGrid when compared within scenarios including a) Goal in Field of View and b) Trigger in Field of View. The distribution and medians of both scenarios suggest that the particular neurons are slightly affected by the presence of the trigger in the field of view, as compared to when goal is in field of view. This is further backed up by the Mann-Whitney U-test, which calculated the neurons activations to be statistically significant}
\end{figure}

\end{document}